\documentclass{article}

\usepackage[preprint]{neurips_2026}

\usepackage[utf8]{inputenc}
\usepackage[T1]{fontenc}
\usepackage{hyperref}
\usepackage{url}
\usepackage{booktabs}
\usepackage{amsfonts}
\usepackage{nicefrac}
\usepackage{microtype}
\usepackage{graphicx}
\usepackage{amsmath}
\usepackage{amssymb}
\usepackage{algorithm}
\usepackage{algorithmic}
\usepackage{xcolor}
\usepackage{multirow}
\usepackage{textcomp}

\usepackage{enumitem}

\DeclareUnicodeCharacter{202F}{\,}
\DeclareUnicodeCharacter{223C}{$\sim$}

\graphicspath{{figures/}}

\title{iPhoneBlur: A Difficulty-Stratified Benchmark for Consumer Device Motion Deblurring}

\author{%
  Abdullah Al Shafi\textsuperscript{1}, Kazi Saeed Alam\textsuperscript{1} \\[1ex]
  \textsuperscript{1}Department of Computer Science and Engineering \\
  Khulna University of Engineering \& Technology, Bangladesh \\
  \texttt{abdullah.shafi99@gmail.com, saeed.alam@cse.kuet.ac.bd}
}

\begin{document}

\maketitle

\begin{abstract}
Motion blur restoration on consumer mobile devices is typically evaluated using aggregate metrics that obscure performance variation across blur difficulty, masking model behavior under real deployment conditions. This work introduces iPhoneBlur, a difficulty-stratified benchmark of 7,400 image pairs synthesized from high-framerate iPhone 17 Pro videos captured in diverse real-world scenarios. Samples are partitioned into Easy, Medium, and Hard categories through PSNR-guided adaptive temporal windowing, with stratification validated by monotonic $2.2\times$ increase in optical flow magnitude across tiers. Each sample includes comprehensive metadata enabling investigation of ISP-aware and difficulty-adaptive restoration strategies. Spectral analysis confirms synthesized blur exhibits high-frequency suppression patterns consistent with authentic motion degradation. Evaluation of six architectures reveals consistent 7--9~dB performance degradation from Easy to Hard subsets, a substantial gap entirely hidden by aggregate reporting. The benchmark further exposes a domain gap between professional and consumer cameras which targeted fine-tuning substantially recovers. By coupling difficulty stratification with deployment-critical metadata, iPhoneBlur enables systematic assessment of model reliability and failure modes for resource-constrained edge systems.
\end{abstract}

\section{Introduction}

Motion blur degrades billions of smartphone photos daily, yet state-of-the-art deblurring methods are developed and evaluated almost exclusively on professional camera equipment. This domain mismatch creates a critical deployment gap: while methods achieve strong performance on high-end action cameras and DSLR rigs~\cite{nah2017deep,rim2020real,shen2019human}, models trained on professional datasets experience severe degradation when applied to consumer mobile captures, with performance dropping by over 7~dB. Beyond device differences, existing benchmarks report only aggregate metrics, concealing performance variation across blur severity—variation that determines whether edge devices can restore images on-device or must offload to cloud infrastructure~\cite{sandler2018mobilenetv2,figurnov2017spatially}.

This work introduces iPhoneBlur, the first difficulty-stratified benchmark for consumer device motion deblurring. The dataset provides 7,400 blur-sharp image pairs captured from diverse real-world scenarios using an iPhone 17 Pro at high framerate (177--240~fps), then processed through PSNR-guided adaptive synthesis to ensure controlled difficulty partitioning. Unlike existing benchmarks that provide only aggregate test sets, iPhoneBlur categorizes samples into Easy, Medium, and Hard tiers validated through independent physical motion analysis. Each sample includes comprehensive metadata—ISP characteristics, optical flow magnitude, noise estimates, and synthesis parameters—enabling research directions absent from prior benchmarks: ISP-aware restoration modeling camera-specific artifacts, difficulty prediction for adaptive inference, and systematic domain transfer from professional to consumer captures.

\begin{figure}[t]
\centering
\includegraphics[width=0.95\textwidth]{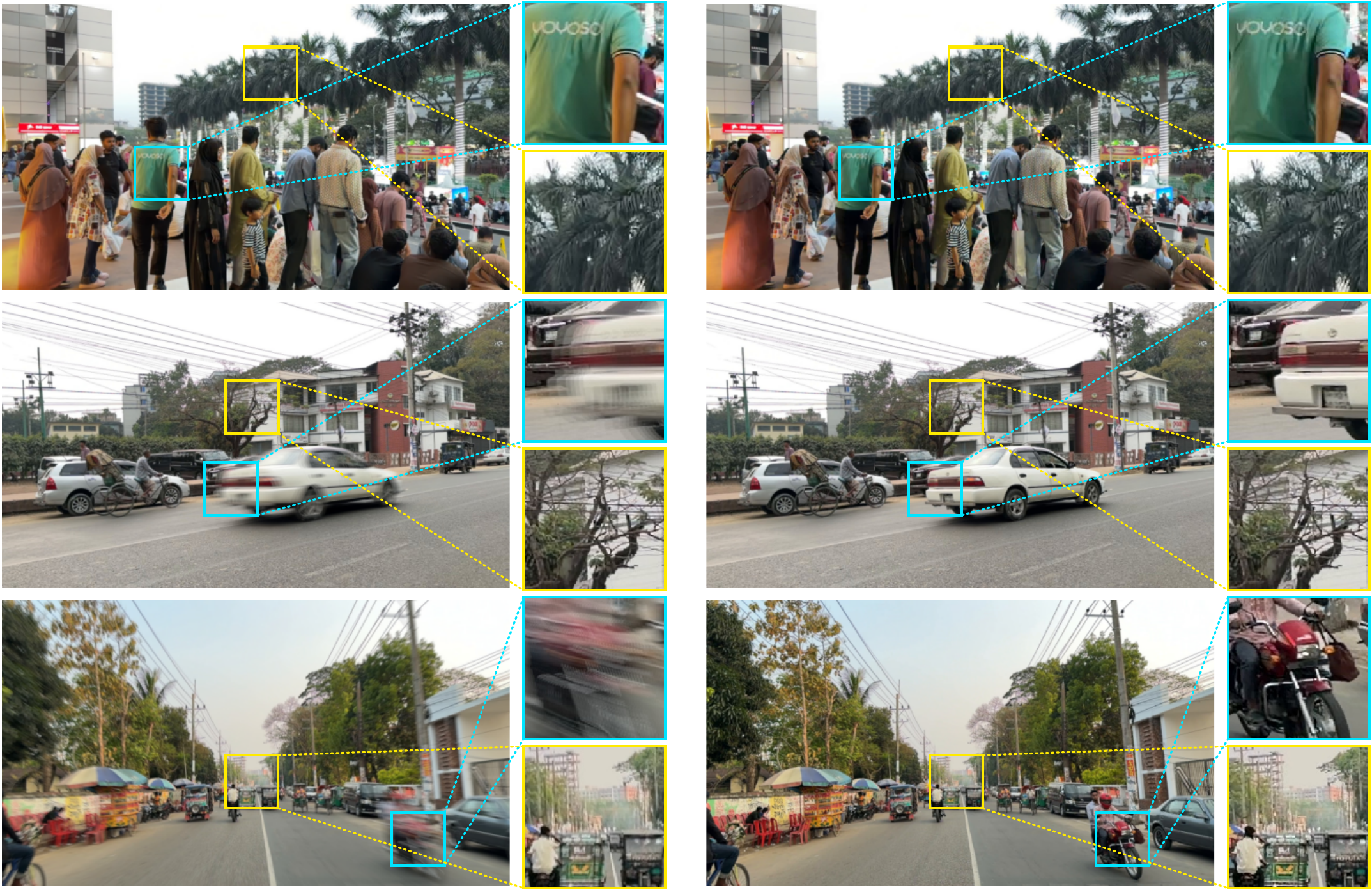}
\caption{Representative blur-sharp pairs across difficulty levels. Easy samples retain fine detail, Medium show moderate motion streaks, and Hard suffer severe degradation. Existing benchmarks conceal such failure modes through aggregate reporting.}
\label{fig:samples}
\end{figure}

Difficulty stratification is validated through physical scene properties and synthesis quality analysis. Optical flow magnitude exhibits monotonic increase across difficulty tiers (2.2$\times$ from Easy to Hard), with strong negative correlation to PSNR ($\rho=-0.41$, $p<10^{-294}$), confirming stratification reflects actual motion dynamics rather than arbitrary thresholding. Synthesis quality is verified through high-frequency suppression analysis: linearized temporal averaging produces Cohen's $d=2.32$, approaching authentic blur characteristics observed in RealBlur-J ($d=2.24$) and matching established synthetic benchmarks. Comprehensive baseline evaluation across six diverse architectures exposes consistent 7--9~dB performance degradation from Easy to Hard samples, revealing robustness limitations entirely masked when only aggregate scores are reported. The benchmark further demonstrates that targeted fine-tuning substantially narrows the professional-to-consumer domain gap, validating its utility for systematic adaptation research.

The benchmark contributes: \textbf{(1)} validated difficulty stratification enabling fine-grained performance analysis essential for deployment decisions, exposing architecture-agnostic failure patterns hidden by aggregate metrics; \textbf{(2)} comprehensive per-sample metadata supporting ISP-aware methods, difficulty-adaptive inference, and computational routing decisions; \textbf{(3)} rigorous synthesis validation through physics-based temporal integration analysis with cross-dataset consistency; \textbf{(4)} consumer device focus bridging the professional-to-mobile gap with demonstrated adaptation pathways. 

\section{Related Work}

\subsection{Motion Deblurring Benchmarks}

Existing benchmarks provide substantial scale but lack difficulty characterization. GoPro~\cite{nah2017deep} established linearized temporal averaging for 3,214 synthetic pairs from 240fps action cameras. REDS~\cite{nah2019ntire} scaled to 240 video sequences (24,000 frames) for multi-frame restoration. HIDE~\cite{shen2019human} focused on human motion with 2,025 test pairs. RealBlur~\cite{rim2020real} bypassed synthesis entirely, capturing authentic blur via beam-splitter optical setup (4,738 pairs). However, these datasets rely on action cameras or high-end mirrorless equipment with image signal processing (ISP) characteristics fundamentally different from modern consumer mobile devices. Furthermore, while some restoration methods implicitly handle varying blur levels through multi-scale architectures, no existing benchmark provides validated stratification grounded in physical scene motion for systematic evaluation. In contrast, this work introduces validated physical-motion-grounded stratification with comprehensive per-sample metadata, addressing the gap in deployment-critical difficulty-based evaluation.

\subsection{Synthesis Quality and Validation}

Realistic blur synthesis requires modeling camera sensor physics and ISP pipelines. Nah et al. distinguished naive sRGB averaging—which produces gamma-encoding artifacts—from linearized averaging that approximates sensor temporal integration. Subsequent synthetic benchmarks adopt linearized synthesis approaches validated primarily through perceptual metrics (SSIM~\cite{wang2004image}, LPIPS~\cite{zhang2018unreasonable}) and visual inspection. ~\cite{brooks2019unprocessing} formalized camera unprocessing, emphasizing linear-domain operations for authentic image formation modeling. This work extends validation beyond perceptual metrics through quantitative high-frequency suppression analysis, providing physics-based consistency verification that demonstrates statistical alignment with authentic camera blur characteristics.

\section{Dataset Construction Methodology}

\subsection{Design Principles and Pipeline Overview}

iPhoneBlur construction addresses three gaps in existing benchmarks: lack of explicit difficulty categorization preventing deployment-critical performance analysis~\cite{figurnov2017spatially}, stratification methods relying on post-hoc grouping rather than validated physical scene properties, and synthesis quality assessed via perceptual metrics alone without quantitative consistency checks.

\begin{figure}[t]
\centering
\includegraphics[width=0.95\textwidth]{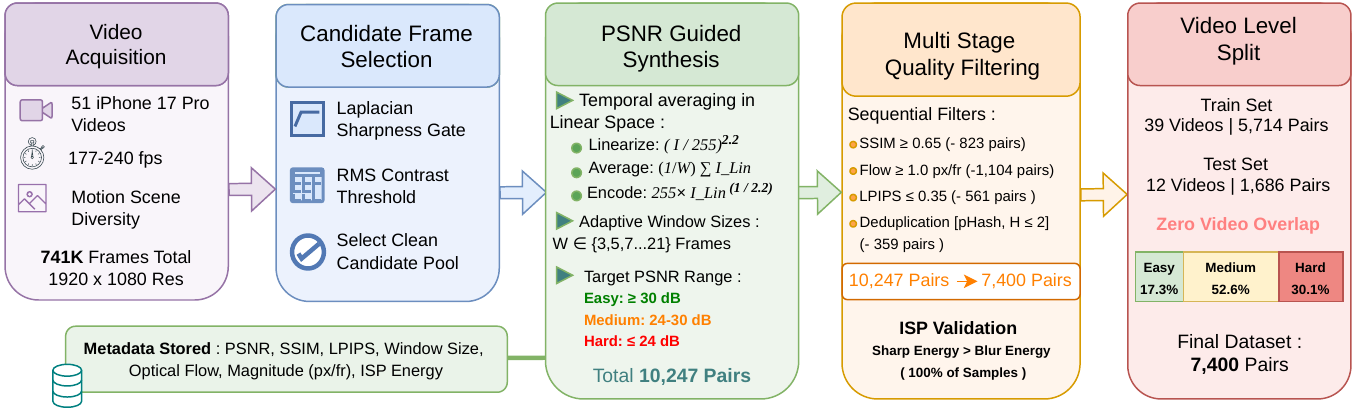}
\caption{\textbf{Dataset construction pipeline.} Five-stage process: candidate selection from 741K frames, $\gamma$-linearized temporal synthesis with adaptive windowing (Eq.~\ref{eq:linearization}), sequential quality filtering, and video-level split yielding 7,400 difficulty-stratified pairs.}
\label{fig:pipeline}
\end{figure}

Source material comprises 51 high-framerate videos (iPhone 17 Pro, 177--240~fps, 1920$\times$1080, $\sim$741K frames). High temporal resolution (4--6~ms inter-frame spacing) enables precise blur duration control through adaptive window selection. Content spans indoor/outdoor environments, natural settings, urban scenes with vehicular/pedestrian motion, human activities, and dynamic phenomena. Candidate sharp frames identified via multi-stage filtering: sharpness (Laplacian variance~\cite{pan2016blind}), contrast (RMS luminance). Sampling stride adapts to video length (15 frames for videos $<$800 frames, 30 frames otherwise) maintaining candidate pool density. Figure~\ref{fig:pipeline} illustrates the complete five-stage construction pipeline.

\subsection{PSNR-Guided Linearized Synthesis with Adaptive Window Selection}

Given candidate sharp frame $I_{t_c}$, blur is synthesized by temporal averaging over
a window $W$ centred at $t_c$. The window size adapts to the target PSNR range
$[P_{\min}, P_{\max}]$ of the corresponding difficulty tier (Easy/Medium/Hard).
This adaptive selection—unlike the fixed‑window approaches of GoPro (11 frames) and
HIDE (13 frames)~\cite{nah2017deep,shen2019human}—achieves the target PSNR through
physically realistic temporal integration regardless of scene motion, thereby preventing
spurious difficulty‑content correlation. Algorithm~\ref{alg:synthesis} enumerates
odd‑sized windows ($3, 5, 7, \dots, 21$ frames; odd count ensures symmetric temporal
context around $t_c$) and selects the one closest to the target.

\begin{algorithm}[t]
\caption{PSNR-Guided Linearized Blur Synthesis}
\label{alg:synthesis}
\begin{algorithmic}[1]
\REQUIRE Video $\{I_t\}_{t=1}^T$, target PSNR range $[P_{\min}, P_{\max}]$
\ENSURE Blur-sharp pair $(B, S)$ with $\text{PSNR}(B, S) \in [P_{\min}, P_{\max}]$
\STATE Select center frame $t_c$ from candidate set; set $S \leftarrow I_{t_c}$
\STATE Initialize $W_{\text{best}} \leftarrow \text{NULL}$, $\epsilon_{\min} \leftarrow \infty$
\FOR{$W \in \{3,5,7,\ldots,21\}$}
    \STATE Compute window $[t_s, t_e] \leftarrow [t_c - \lfloor W/2 \rfloor, t_c + \lfloor W/2 \rfloor]$
    \STATE Linearize \& average: $B_{\text{lin}} \leftarrow \frac{1}{W}\sum_{t=t_s}^{t_e} (I_t / 255)^{2.2}$
    \STATE Convert to sRGB: $B_{\text{temp}} \leftarrow 255 \times (B_{\text{lin}})^{1/2.2}$
    \STATE Compute $p \leftarrow \text{PSNR}(B_{\text{temp}}, S)$
    \STATE Compute $\epsilon \leftarrow \min(|p - P_{\min}|, |p - P_{\max}|)$
    \IF{$\epsilon < \epsilon_{\min}$}
        \STATE $W_{\text{best}} \leftarrow W$; $B \leftarrow B_{\text{temp}}$; $\epsilon_{\min} \leftarrow \epsilon$
    \ENDIF
\ENDFOR
\RETURN $(B, S)$ with metadata $(W_{\text{best}}, \text{PSNR}(B, S))$
\end{algorithmic}
\end{algorithm}

Linearization applies inverse sRGB gamma correction ($\gamma=2.2$, sRGB standard):
\begin{equation}
I_{\text{lin}} = \left(\frac{I_{\text{sRGB}}}{255}\right)^{\gamma},\qquad \gamma = 2.2.
\label{eq:linearization}
\end{equation}
This $\gamma^{-1}$ transformation is applied post‑capture to iPhone video frames, which
are recorded with standard ISP processing (tone mapping, HDR, stabilization) and then
linearized for synthesis. The $\gamma=2.2$ power‑law approximation follows established
practice~\cite{nah2017deep,shen2019human}; the exact sRGB piecewise transfer differs
negligibly near black levels and does not affect mid‑tone blur synthesis. The
transformation is essential because camera sensors integrate linear light, and averaging
gamma‑encoded values would produce physically incorrect blur~\cite{brooks2019unprocessing}.
Using Apple’s full production ISP and subsequently linearizing for synthesis ensures the
benchmark reflects real consumer photography conditions, prioritising ecological validity
for deployment. The adaptive search accommodates motion diversity: nearly static scenes
require long windows (e.g., 21 frames yielding 24~dB), whereas dynamic scenes reach the
same PSNR with short windows (e.g., 3 frames). In practice, Easy‑tier samples
predominantly select short windows (3--7 frames), producing mild blur that can be almost
perfectly inverted, consistent with the high PSNR observed.

\subsection{Difficulty Stratification and Quality Filtering}

Difficulty tiers defined by PSNR thresholds corresponding to perceptual degradation boundaries~\cite{hore2010image}: Easy ($\geq$30~dB, just-noticeable-difference threshold, minimal artifacts), Medium (24--30~dB, moderate degradation), Hard ($<$24~dB, substantial degradation). To prevent spurious difficulty-video correlation, construction enforces cross-tier contribution: 40 of 51 source videos (78\%) contribute samples to all three tiers.

Multi-stage quality filtering sequentially applies four criteria to remove synthesis artifacts: minimum SSIM 0.65 ensuring structural preservation, minimum optical flow 1.0~px/frame for Medium/Hard preventing near-static blur, maximum LPIPS 0.35 enforcing perceptual similarity, and near-duplicate removal via perceptual hashing (pHash Hamming distance $\leq$ 2). Sequential application reduces 10,247 candidate pairs to 7,400 final pairs (removing 823, 1,104, 561, and 359 pairs respectively). All samples pass ISP validation: sharp high-frequency energy exceeds blur energy (100\% of pairs, detailed analysis in Section 4.2). Final difficulty distribution: 17.3\% Easy, 52.6\% Medium, 30.1\% Hard.

Video-level train-test split (standard 75/25 ratio~\cite{nah2017deep}) assigns complete videos to training (39 videos, 5,714 pairs) or testing (12 videos, 1,686 pairs), preventing data leakage while maintaining similar difficulty distributions (Easy: 17.0\% train vs 18.3\% test).

Each sample includes comprehensive metadata absent from existing benchmarks. Recorded attributes span three categories: quality metrics (PSNR, SSIM, LPIPS enabling post-hoc filtering or weighted training), physical scene properties (Farnebäck optical flow~\cite{farneback2003two} magnitude 6.0--13.2~px/frame enabling difficulty prediction, Laplacian sharpness variance, RMS contrast), and ISP artifact measurements (sharp/blur high-frequency energy, noise estimates, synthesis parameters including adaptive window size 3--21 frames, temporal bounds, target/achieved PSNR). This metadata enables ISP-aware restoration modeling blur-sharpening interactions during image formation, difficulty prediction for adaptive inference routing, and systematic professional-to-consumer domain adaptation leveraging consumer ISP characteristics. Complete field descriptions provided in Table~\ref{tab:metadata_appendix}.

\section{Dataset Validation}
\label{sec:validation}

\subsection{Physical Validation of Difficulty Stratification}

Difficulty stratification reflects physical scene motion dynamics rather than arbitrary thresholding. Optical flow magnitude increases monotonically across difficulty (Easy 6.0$\pm$5.0 $\to$ Medium 11.3$\pm$5.6 $\to$ Hard 13.2$\pm$4.3~px/frame, 2.2$\times$ overall increase) with strong negative correlation to PSNR (Spearman $\rho=-0.41$, $p<10^{-294}$), confirming categories reflect actual motion rather than synthesis artifacts. Non-parametric Kruskal-Wallis test rejects homogeneity hypothesis ($H=967.6$ train, $H=294.7$ test, both $p<10^{-100}$), validating tier separability. Although difficulty tiers are defined via PSNR, the stratification is independently validated: optical flow magnitude (Spearman $\rho=-0.41$), user perceptual ratings ($\rho=0.74$), and high-frequency suppression (Cohen's $d$ monotonic increase across tiers) all confirm that the Easy-to-Hard performance gaps reflect genuine motion degradation rather than metric artifacts. Table~\ref{tab:dataset_stats} provides comprehensive statistics.

\begin{table}[t]
\centering
\caption{Dataset statistics across difficulty levels. Clear quality separation (9.1~dB PSNR span) with validated motion-based stratification (2.2$\times$ optical flow increase, Spearman $\rho=-0.41$, $p<10^{-294}$).}
\label{tab:dataset_stats}
\small
\begin{tabular}{lcccc}
\toprule
 & All & Easy & Medium & Hard \\
\midrule
Samples (train/test) & 7,400 (5714/1686) & 1,277 (997/280) & 3,895 (2942/953) & 2,228 (1775/453) \\
\midrule
PSNR (dB) & 26.2$\pm$3.4 & 31.7$\pm$1.2 & 26.5$\pm$1.6 & 22.6$\pm$0.9 \\
SSIM & 0.850$\pm$0.090 & 0.960$\pm$0.010 & 0.870$\pm$0.050 & 0.750$\pm$0.060 \\
LPIPS & 0.16$\pm$0.08 & 0.05$\pm$0.02 & 0.14$\pm$0.06 & 0.25$\pm$0.06 \\
\midrule
Optical flow (px/frame) & 11.0$\pm$5.7 & 6.0$\pm$5.0 & 11.3$\pm$5.6 & 13.2$\pm$4.3 \\
Sharpness (Var($\nabla^2$)) & 2137$\pm$622 & 2289$\pm$732 & 2113$\pm$577 & 2091$\pm$633 \\
Contrast (RMS) & 0.267$\pm$0.023 & 0.264$\pm$0.028 & 0.269$\pm$0.021 & 0.264$\pm$0.023 \\
\bottomrule
\end{tabular}
\end{table}

User study validates perceptual alignment: 15 participants rated 30 images on 5-point severity scale (1=imperceptible blur, 5=severe degradation), yielding mean ratings 2.0$\pm$0.8 (Easy), 3.2$\pm$0.9 (Medium), 4.1$\pm$0.7 (Hard), with Spearman correlation $\rho=0.74$ ($p<0.001$) confirming difficulty-perception correspondence.

\begin{figure}[t]
\centering
\begin{minipage}[b]{0.48\textwidth}
\includegraphics[width=\textwidth]{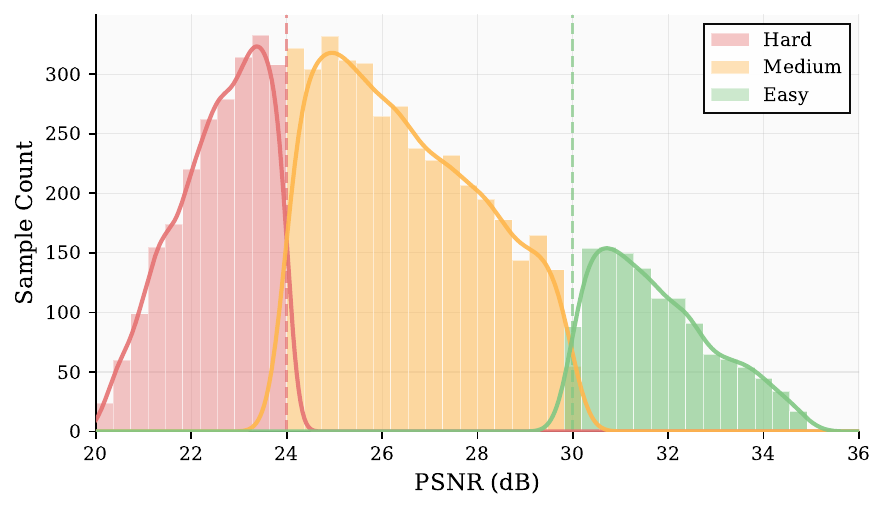}
\end{minipage}
\hfill
\begin{minipage}[b]{0.48\textwidth}
\includegraphics[width=\textwidth]{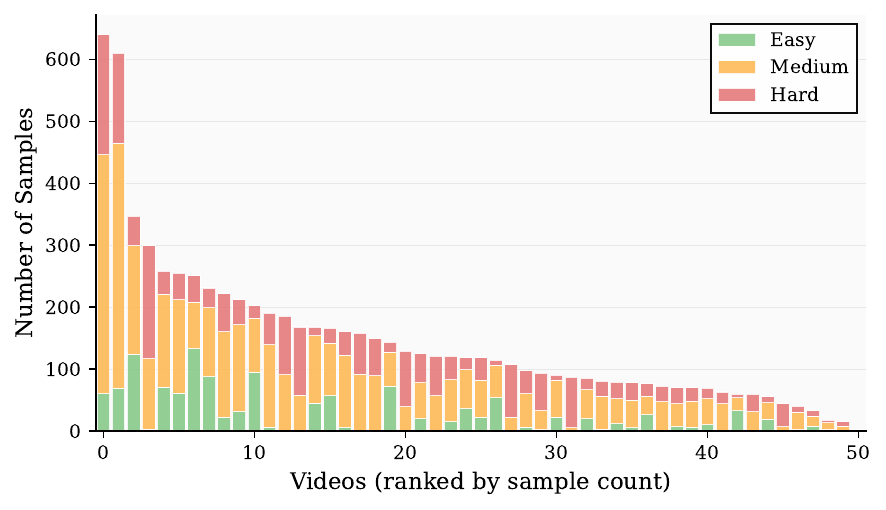}
\end{minipage}
\caption{\textbf{Dataset visualization.} Left: PSNR distribution across difficulty tiers. Right: Video contribution—78\% span all tiers.}
\label{fig:dist}
\end{figure}

\subsection{Synthesis Quality Validation and Benchmark Positioning}

Realistic motion blur synthesis exhibits systematic high-frequency suppression—temporal integration signature. Validation employs mean absolute Laplacian energy $E(I) = \frac{1}{|\Omega|}\sum_{(x,y)}|\nabla^2 L|$ where $L$ is ITU-R BT.601 luminance~\cite{wang2004image}. Per-sample motion computed as average Farnebäck optical flow magnitude over synthesis window:
\begin{equation}
m_i = \frac{1}{W_i}\sum_{t=t_s}^{t_e} \mathbb{E}_{(x,y)}\bigl[\|\mathbf{v}_t(x,y)\|\bigr]
\label{eq:motion_magnitude}
\end{equation}
where $\mathbf{v}_t$ is dense optical flow between consecutive frames (px/frame).

\begin{figure}[t]
\centering
\begin{minipage}[b]{0.48\textwidth}
\includegraphics[width=\textwidth]{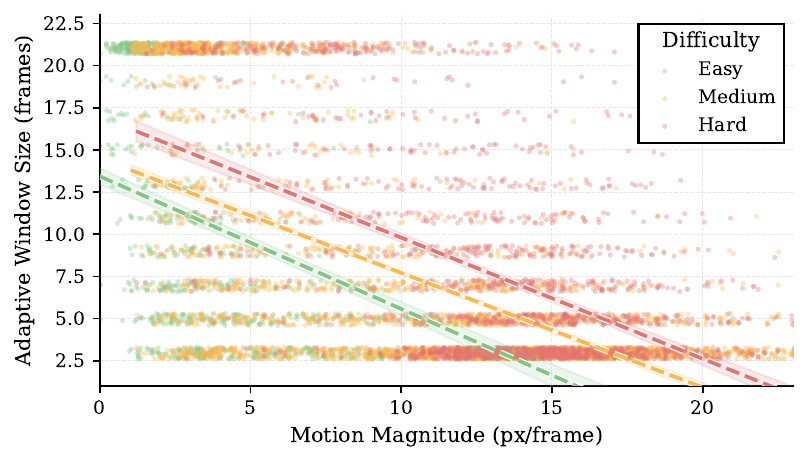}
\end{minipage}
\hfill
\begin{minipage}[b]{0.48\textwidth}
\includegraphics[width=\textwidth]{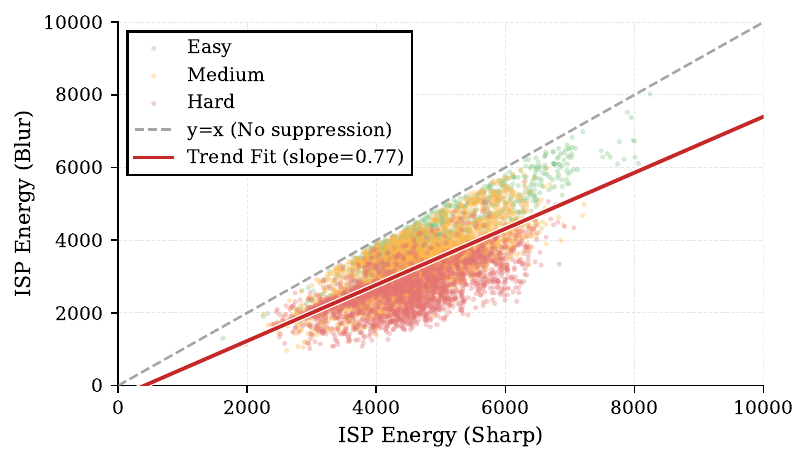}
\end{minipage}
\caption{\textbf{Motion-difficulty relationship and synthesis validation.} Left: Adaptive window size vs optical flow—high motion requires shorter windows to achieve target PSNR. Right: ISP energy scatter—100\% samples exhibit suppression (below diagonal), slope 0.77 confirms 23\% high-frequency reduction.}
\label{fig:motion_isp}
\end{figure}

High-frequency suppression analysis (Figure~\ref{fig:motion_isp}, right panel) confirms synthesis quality: iPhoneBlur exhibits Cohen's $d$ ranging 1.84--2.64 across difficulty levels (overall mean 2.32), approaching RealBlur-J authentic blur ($d=2.24$) and established synthetic benchmarks (GoPro $d=2.81$, HIDE $d=2.79$, REDS $d=2.45$). High-frequency energy (measured as mean absolute Laplacian) exhibits universal suppression across 100\% samples with linear regression slope 0.77, demonstrating 23\% reduction consistent with temporal integration physics. Monotonic strengthening across difficulty (Easy 2.10 $\to$ Hard 2.64, training split) validates adaptive window methodology. Per-difficulty breakdown provided in Table~\ref{tab:cohens_appendix}. Threshold generalization validated across datasets: applying identical PSNR thresholds
to GoPro (1,111 test pairs), HIDE (2,025 test pairs), and RealBlur‑J (980 pairs) yields
consistent Medium concentration (39--60\%), confirming thresholds capture intrinsic
difficulty rather than dataset‑specific artifacts.

Table~\ref{tab:benchmark_comparison} positions iPhoneBlur against established benchmarks. While GoPro and HIDE achieve strong synthesis quality, they lack explicit difficulty stratification and employ action camera capture equipment (GoPro Hero 240fps). REDS offers large scale (24K frames) with video-level split but no difficulty tiers. RealBlur bypasses synthesis entirely through beam-splitter optical systems but provides no stratification. 

\begin{table}[t]
\centering
\caption{Benchmark comparison. iPhoneBlur uniquely combines difficulty stratification (3 validated levels), consumer device capture, and comprehensive per-sample metadata enabling ISP-aware restoration and difficulty-adaptive inference.}
\label{tab:benchmark_comparison}
\small
\begin{tabular}{lcccccc}
\toprule
Dataset & Pairs & Device & Stratified & Cohen's $|d|$ & Split & Metadata \\
\midrule
GoPro & 3,214 & GoPro Hero (240fps) & No & 2.81 & Sample & No \\
HIDE & 2,025 & GoPro Hero (240fps) & No & 2.79 & Sample & No \\
RealBlur-J & 980 & Canon DSLR & No & 2.24 & Sample & No \\
RealBlur-R & 980 & Nikon DSLR & No & 0.68 & Sample & No \\
REDS & 24,000 & GoPro Hero (120fps) & No & 2.45 & Video & No \\
\midrule
\textbf{iPhoneBlur} & \textbf{7,400} & \textbf{iPhone 17 Pro (240fps)} & \textbf{3 levels} & \textbf{2.32} & \textbf{Video} & \textbf{Yes} \\
\bottomrule
\end{tabular}
\end{table}

\section{Baseline Evaluation and Analysis}
\label{sec:experiments}

\subsection{Experimental Configuration}

Six state-of-the-art methods representing distinct architectural paradigms are evaluated:
the CNN‑based NAFNet~\cite{chen2022simple},
the hierarchical HINet~\cite{chen2021hinet},
the transformer‑based Restormer~\cite{zamir2022restormer},
the multi‑scale MIMO‑UNet~\cite{cho2021rethinking},
the instruction‑guided Instruct‑IR~\cite{conde2024instructir}, and
the frequency‑domain FFTformer~\cite{kong2022efficient}.
By spanning a comprehensive range of computational strategies—from spatial‑domain CNNs
to frequency‑domain transformers—the evaluation ensures that benchmarks capture
fundamental dataset characteristics rather than architecture‑specific inductive biases.
All cross‑domain assessments use official GoPro‑pretrained weights to maintain a
standardized baseline. Detailed fine‑tuning configurations are provided in
Appendix~\ref{app:training}; all reported values represent means over three random seeds.

\subsection{Cross-Domain Zero-Shot Performance}

Table~\ref{tab:cross_domain} presents zero-shot performance using GoPro-pretrained models.
While achieving 31.60--34.0~dB on GoPro, 28.2--31.6~dB on HIDE, and 26.2--35.6~dB on
RealBlur datasets, all methods degrade to 25.2--26.9~dB on iPhoneBlur.
Domain gap varies by architecture: NAFNet and FFTformer exhibit 7.1~dB degradation,
while HINet and Restormer show the smallest drops at 6.1~dB. Instruct‑IR drops 6.4~dB.
Average degradation is 6.6~dB, substantially exceeding typical 2--3~dB drops in
professional camera cross-evaluation, confirming consumer device blur presents distinct
challenges beyond motion statistics or capture conditions.

\begin{table}[t]
\centering
\caption{Zero-shot cross-domain performance (PSNR/SSIM) using GoPro-pretrained models.
iPhoneBlur exhibits 6.1--7.1~dB degradation, confirming consumer device blur domain gap.}
\label{tab:cross_domain}
\small
\begin{tabular}{lcccccc}
\toprule
Test Dataset & NAFNet & HINet & Restormer & MIMO-UNet & Instruct-IR & FFTformer \\
\midrule
GoPro        & 33.7/0.950 & 32.9/0.941 & 32.9/0.961 & 32.4/0.932 & 31.60/0.890 & 34.0/0.965 \\
HIDE         & 31.3/0.920 & 30.2/0.915 & 31.2/0.942 & 30.0/0.905 & 28.2/0.870 & 31.6/0.945 \\
RealBlur-J   & 28.5/0.865 & 27.2/0.835 & 29.0/0.880 & 26.9/0.824 & 26.2/0.810 & 29.2/0.885 \\
RealBlur-R   & 35.1/0.942 & 34.2/0.928 & 35.4/0.948 & 34.0/0.920 & 31.9/0.895 & 35.6/0.950 \\
\midrule
iPhoneBlur   & 26.6/0.868 & 26.8/0.872 & 26.8/0.895 & 25.8/0.861 & 25.2/0.842 & 26.9/0.882 \\
\bottomrule
\end{tabular}
\end{table}

\subsection{Stratified Performance Analysis}

Fine-tuning on iPhoneBlur yields substantial improvements (Table~\ref{tab:finetuned}),
with overall performance ranging 29.5--31.2~dB. Stratified evaluation reveals critical
patterns that aggregate-only benchmarks obscure. All methods exhibit substantial
degradation from Easy to Hard samples, ranging 7.2--9.1~dB—NAFNet (31.2~dB overall)
degrades 7.2~dB while Instruct-IR (29.5~dB overall) shows 9.1~dB drop. This substantially
exceeds typical 1--2~dB variation in aggregate test sets. Easy-to-Hard degradation
(7.2--9.1~dB) even exceeds cross-domain gap (6.1--7.1~dB) from zero-shot evaluation—
demonstrating motion severity poses fundamentally greater challenge than domain
adaptation. This suggests that severe motion blur may constitute a harder ``concept
shift'' than camera pipeline differences alone.

Performance patterns across difficulty strata reveal architectural dependencies. NAFNet
achieves highest overall PSNR (31.2~dB), with FFTformer (31.0~dB) and Restormer (30.9~dB)
competitive. However, Hard sample performance diverges: simpler CNNs (NAFNet 27.9~dB,
HINet 27.1~dB) maintain relative competitiveness, while MIMO-UNet (25.7~dB) and
Instruct-IR (25.4~dB) degrade further despite multi-scale and instruction-guidance
mechanisms. This suggests architectural complexity does not guarantee robustness to
severe motion, with simpler designs potentially offering more stable degradation by
avoiding failure modes when architectural constraints encounter motion deviating from
training assumptions.

Degradation consistency across diverse architectures—simple CNNs (NAFNet), hierarchical
networks (HINet), transformers (Restormer, FFTformer), multi-scale designs (MIMO-UNet),
and instruction-guided methods (Instruct-IR)—exhibits architecture-agnostic standard
deviation of only 0.7~dB, validating that difficulty tiers capture fundamental physical
restoration challenges rather than method-specific biases. Multiple independent validation
signals converge: correlation to optical flow (6.0--13.2~px/frame, Spearman $\rho=-0.41$,
$p<10^{-294}$), human perceptual agreement ($\rho=0.74$, $p<0.001$), monotonic
high-frequency suppression (Cohen's $d$ 2.10$\to$2.64), and architecture-independent
degradation, collectively confirming stratification reflects genuine motion-based
difficulty rather than arbitrary PSNR thresholding.

\begin{table}[t]
\centering
\caption{Fine-tuned performance on iPhoneBlur test set, stratified by difficulty.
Consistent 7.2--9.1~dB degradation across diverse architectures validates difficulty
stratification captures fundamental restoration challenges.}
\label{tab:finetuned}
\small
\setlength{\tabcolsep}{2.8pt}
\begin{tabular}{l@{\hspace{8pt}}l@{\hspace{5pt}}c@{\hspace{3pt}}c@{\hspace{3pt}}c@{\hspace{5pt}}c@{\hspace{3pt}}c@{\hspace{3pt}}c@{\hspace{5pt}}c@{\hspace{3pt}}c@{\hspace{3pt}}c@{\hspace{5pt}}c@{\hspace{3pt}}c@{\hspace{3pt}}c}
\toprule
& & \multicolumn{3}{c}{\textbf{Overall}} & \multicolumn{3}{c}{\textbf{Easy}} & \multicolumn{3}{c}{\textbf{Medium}} & \multicolumn{3}{c}{\textbf{Hard}} \\
\cmidrule(lr){3-5} \cmidrule(lr){6-8} \cmidrule(lr){9-11} \cmidrule(lr){12-14}
\textbf{Method} & & PSNR & SSIM & LPIPS & PSNR & SSIM & LPIPS & PSNR & SSIM & LPIPS & PSNR & SSIM & LPIPS \\
\midrule
NAFNet          && 31.2 & 0.945 & 0.049 & 35.1 & 0.978 & 0.020 & 31.6 & 0.953 & 0.042 & 27.9 & 0.907 & 0.080 \\
\addlinespace[2pt]
HINet           && 30.5 & 0.938 & 0.056 & 34.6 & 0.975 & 0.023 & 30.9 & 0.947 & 0.048 & 27.1 & 0.895 & 0.092 \\
\addlinespace[2pt]
Restormer       && 30.9 & 0.940 & 0.053 & 34.9 & 0.977 & 0.022 & 31.2 & 0.949 & 0.046 & 27.5 & 0.899 & 0.088 \\
\addlinespace[2pt]
MIMO-UNet       && 29.3 & 0.920 & 0.075 & 33.9 & 0.973 & 0.026 & 29.7 & 0.933 & 0.065 & 25.7 & 0.861 & 0.127 \\
\addlinespace[2pt]
Instruct-IR     && 29.5 & 0.924 & 0.071 & 34.5 & 0.973 & 0.019 & 29.9 & 0.935 & 0.063 & 25.4 & 0.869 & 0.118 \\
\addlinespace[2pt]
FFTformer       && 31.0 & 0.943 & 0.050 & 35.0 & 0.977 & 0.021 & 31.3 & 0.951 & 0.043 & 27.7 & 0.903 & 0.083 \\
\bottomrule
\end{tabular}
\vspace{-2mm}
\end{table}

\begin{figure*}[t]
\centering
\small

\definecolor{easygreen}{RGB}{0, 200, 83}
\definecolor{mediumyellow}{RGB}{255, 193, 7}
\definecolor{hardred}{RGB}{255, 23, 68}

\setlength{\fboxsep}{0pt}
\setlength{\fboxrule}{.7pt}

\begin{tabular}{@{}c@{\hspace{0.8mm}}c@{\hspace{0.8mm}}c@{\hspace{0.8mm}}c@{\hspace{0.8mm}}c@{}}
\textbf{Blur} & \textbf{MIMO-UNet} & \textbf{Restormer} & \textbf{NAFNet} & \textbf{Sharp} \\[2pt]

\fcolorbox{easygreen}{white}{\includegraphics[width=0.192\textwidth]{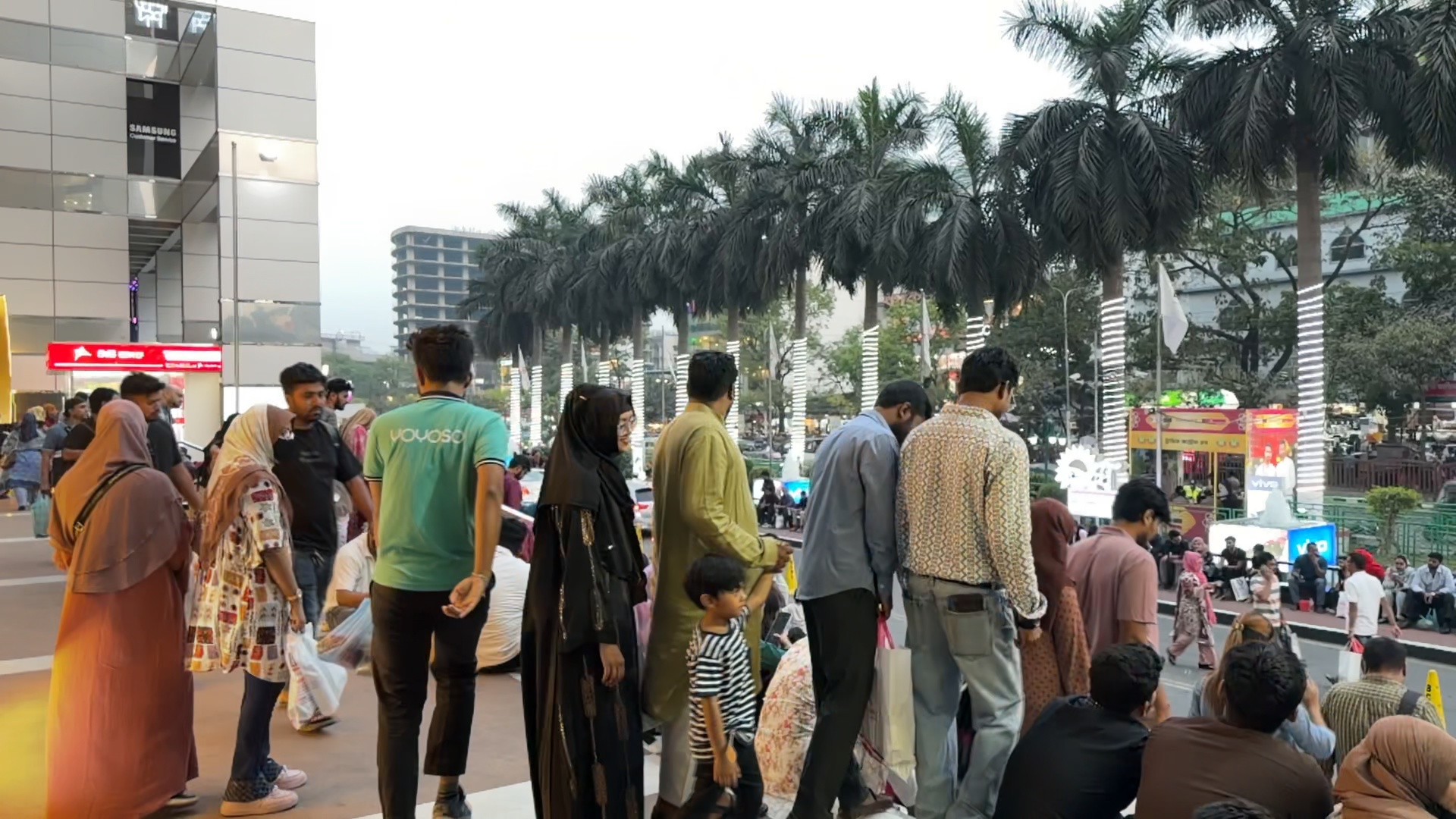}} &
\fcolorbox{easygreen}{white}{\includegraphics[width=0.192\textwidth]{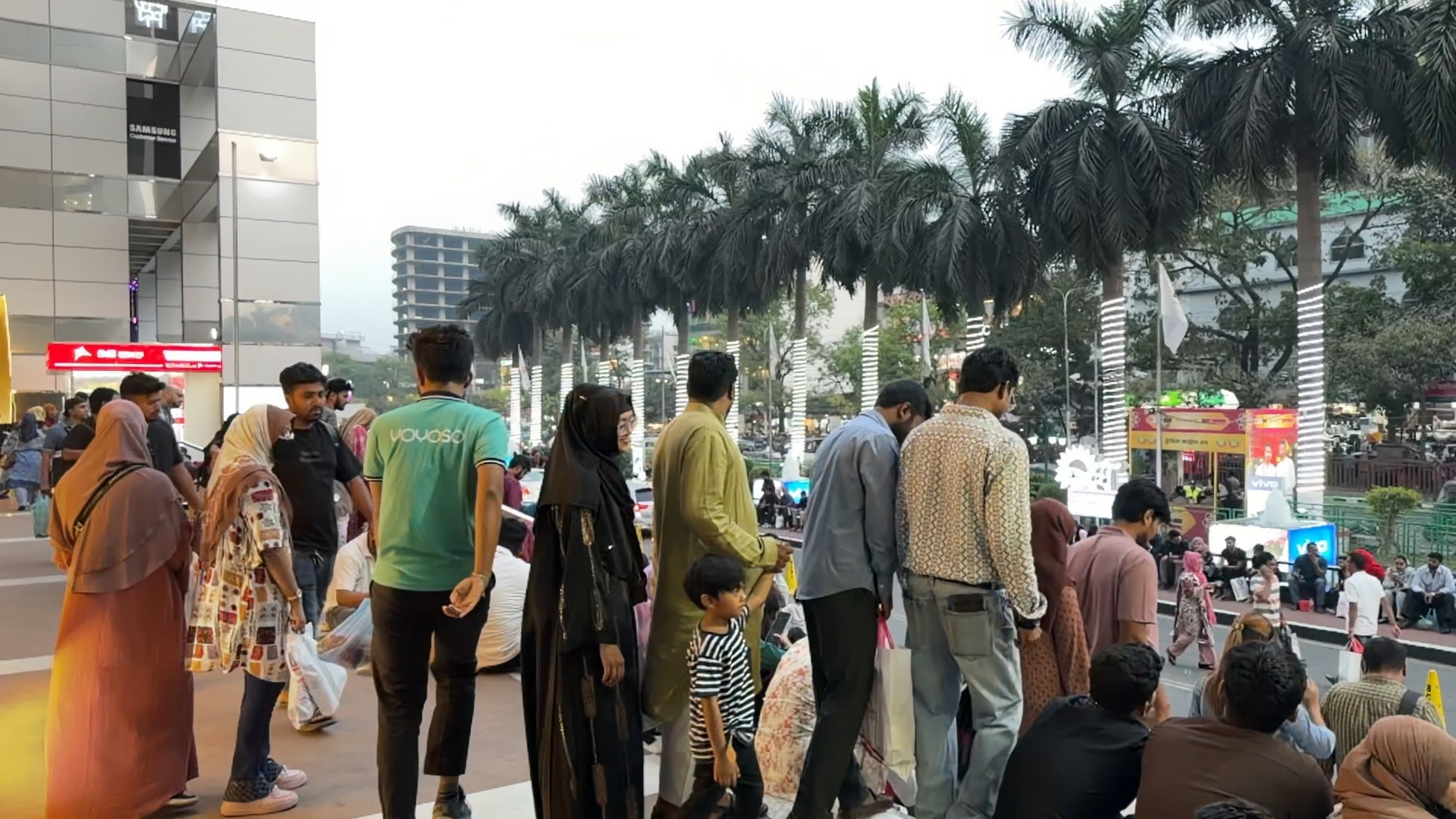}} &
\fcolorbox{easygreen}{white}{\includegraphics[width=0.192\textwidth]{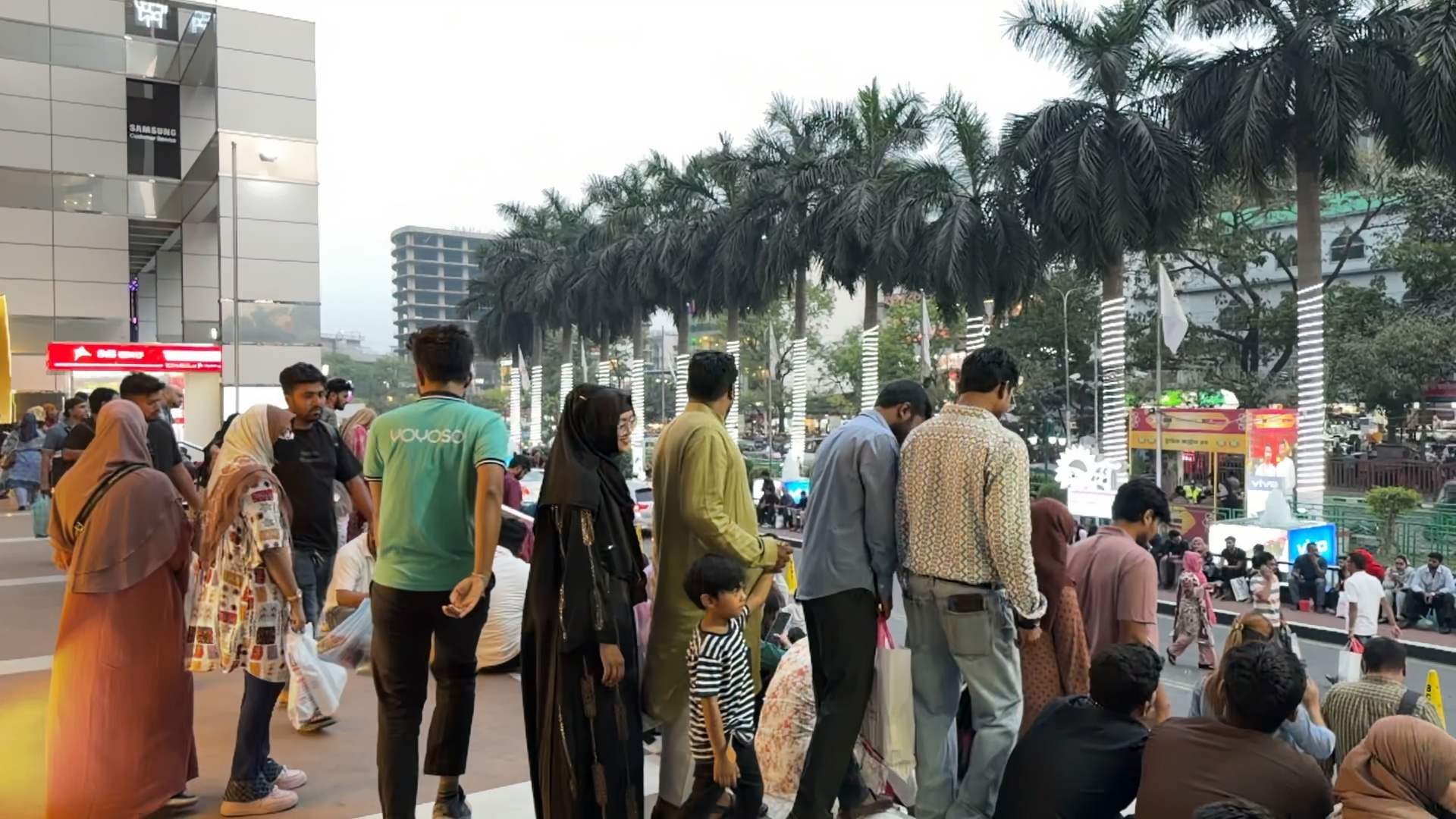}} &
\fcolorbox{easygreen}{white}{\includegraphics[width=0.192\textwidth]{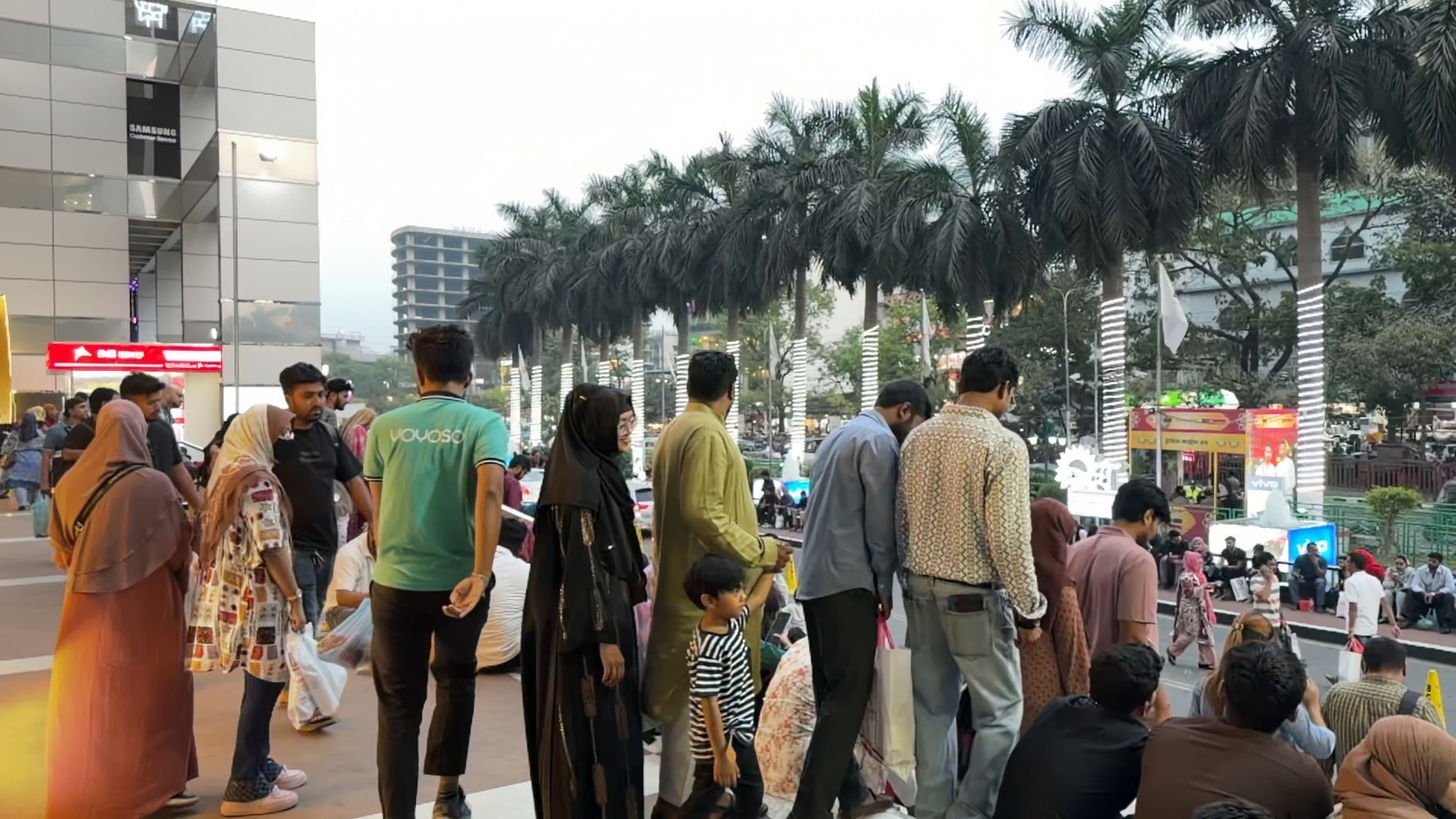}} &
\fcolorbox{easygreen}{white}{\includegraphics[width=0.192\textwidth]{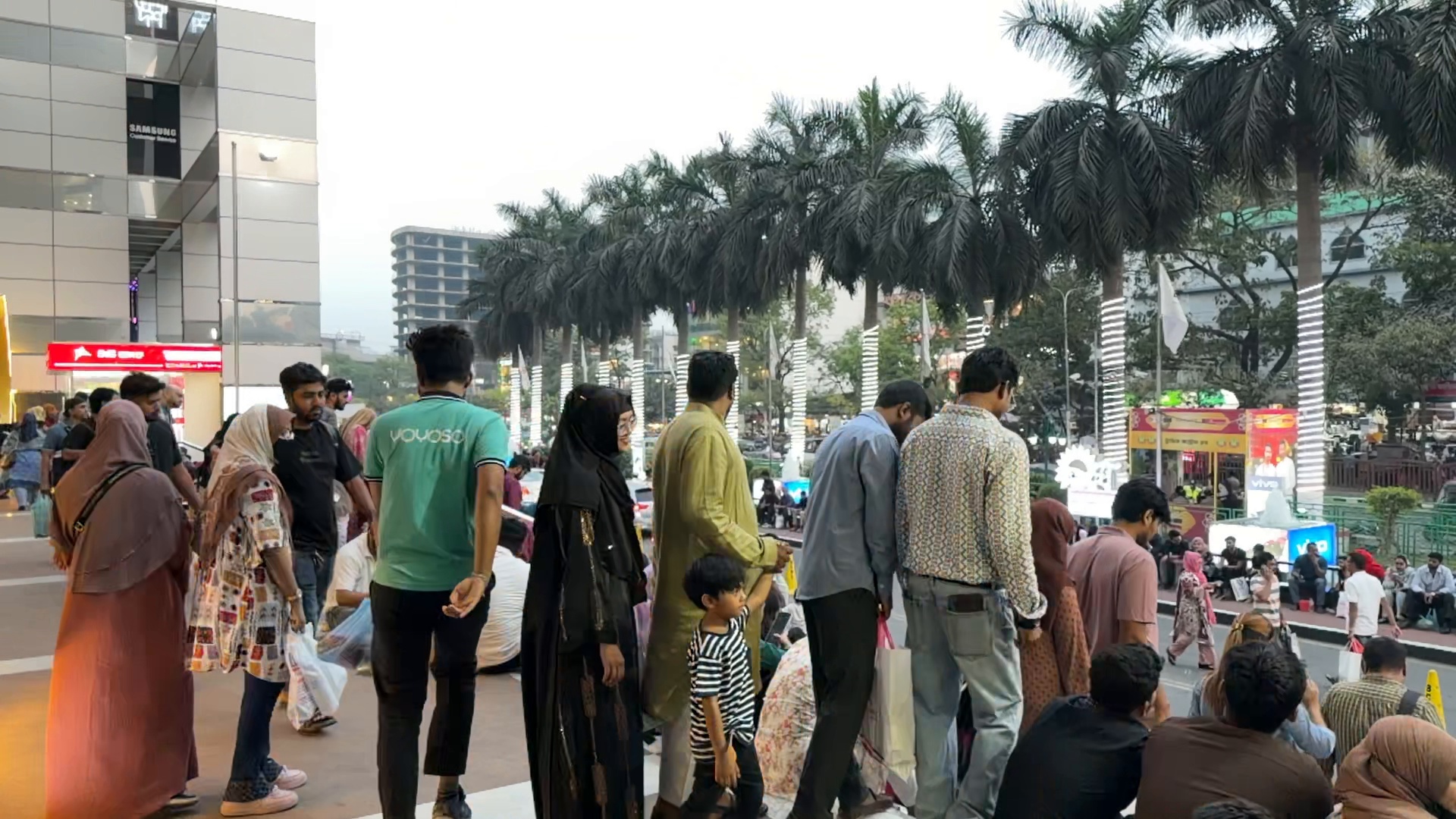}} \\[1.5mm]

\fcolorbox{mediumyellow}{white}{\includegraphics[width=0.192\textwidth]{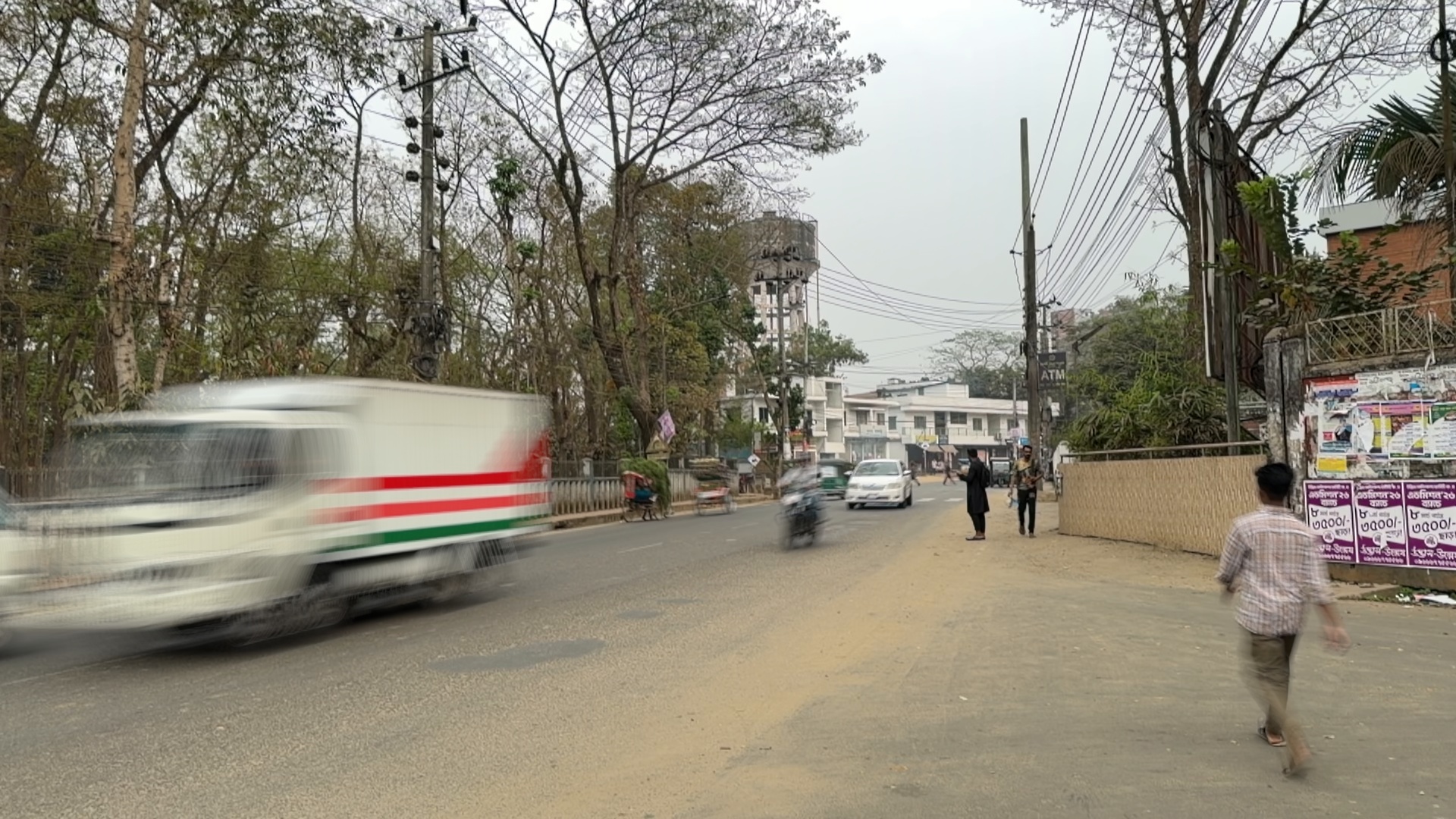}} &
\fcolorbox{mediumyellow}{white}{\includegraphics[width=0.192\textwidth]{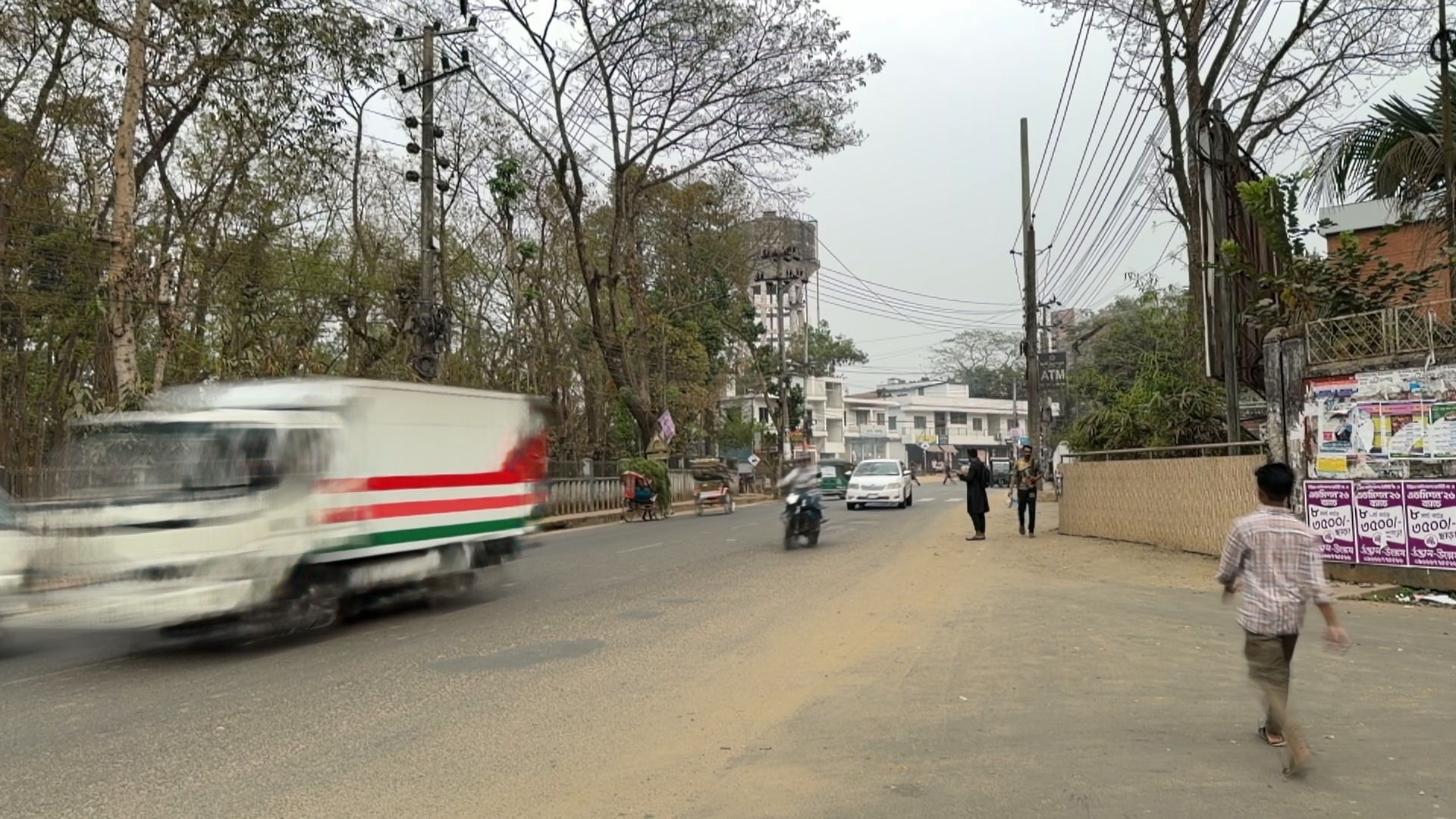}} &
\fcolorbox{mediumyellow}{white}{\includegraphics[width=0.192\textwidth]{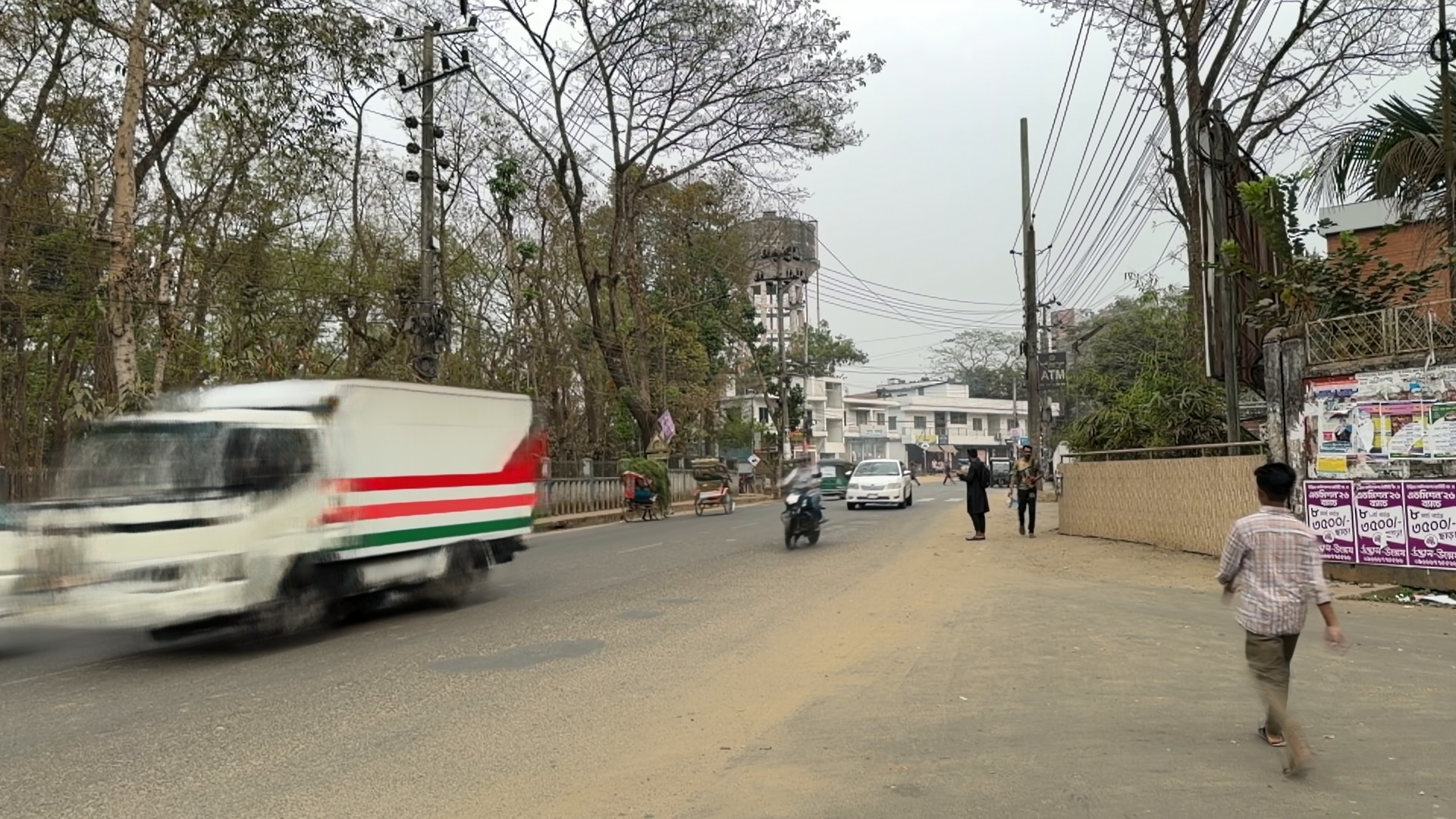}} &
\fcolorbox{mediumyellow}{white}{\includegraphics[width=0.192\textwidth]{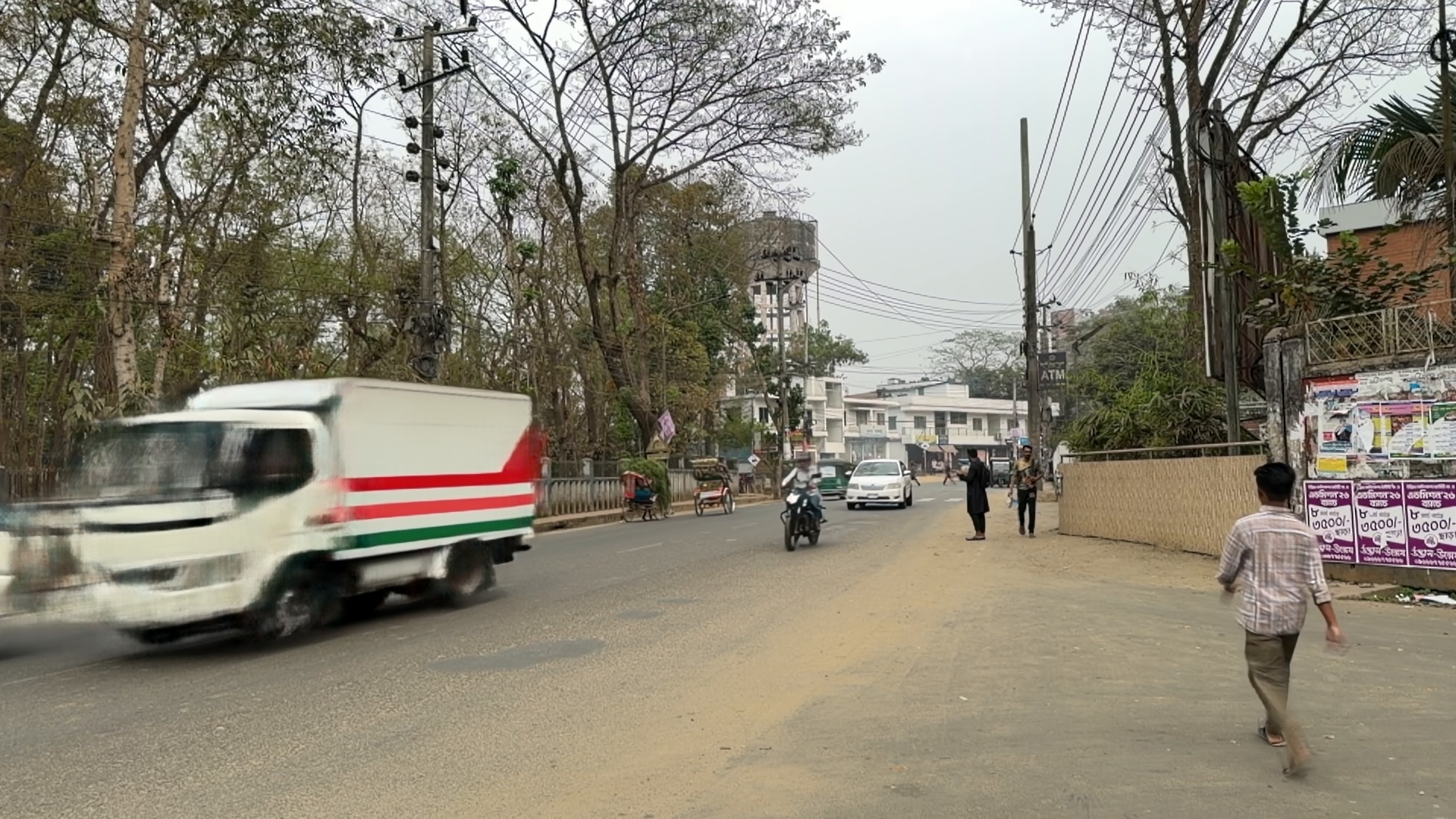}} &
\fcolorbox{mediumyellow}{white}{\includegraphics[width=0.192\textwidth]{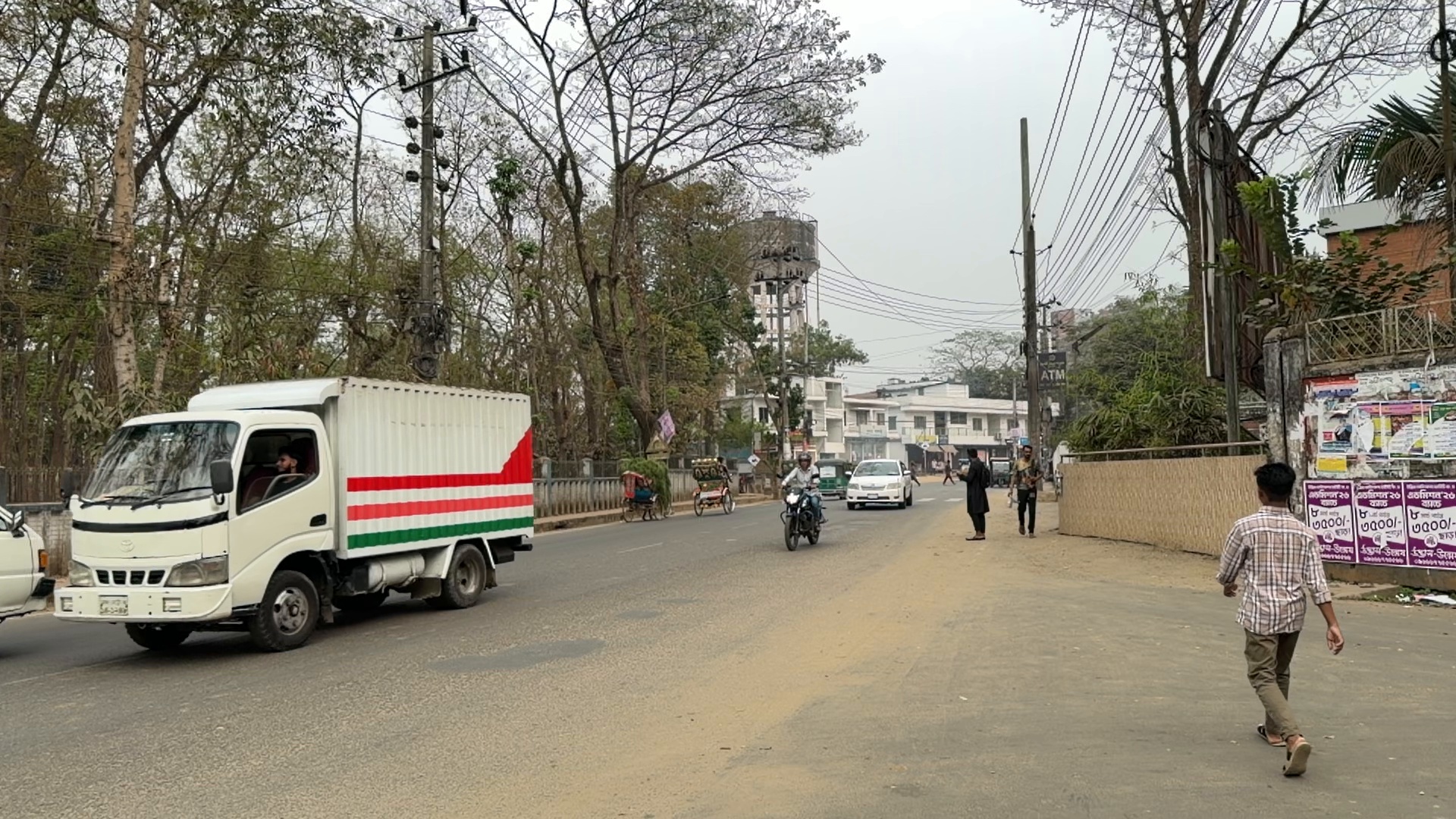}} \\[1.5mm]

\fcolorbox{mediumyellow}{white}{\includegraphics[width=0.192\textwidth]{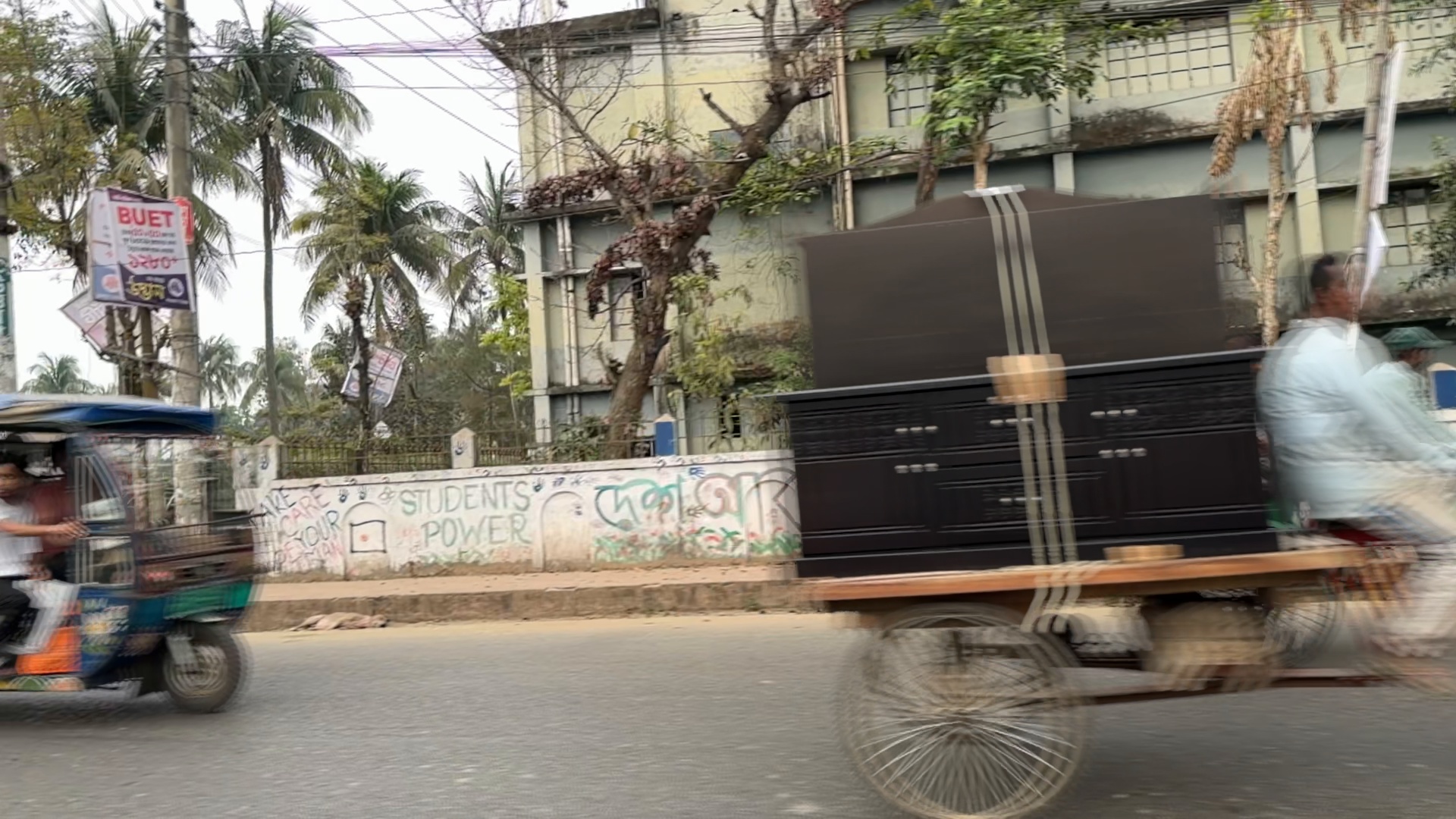}} &
\fcolorbox{mediumyellow}{white}{\includegraphics[width=0.192\textwidth]{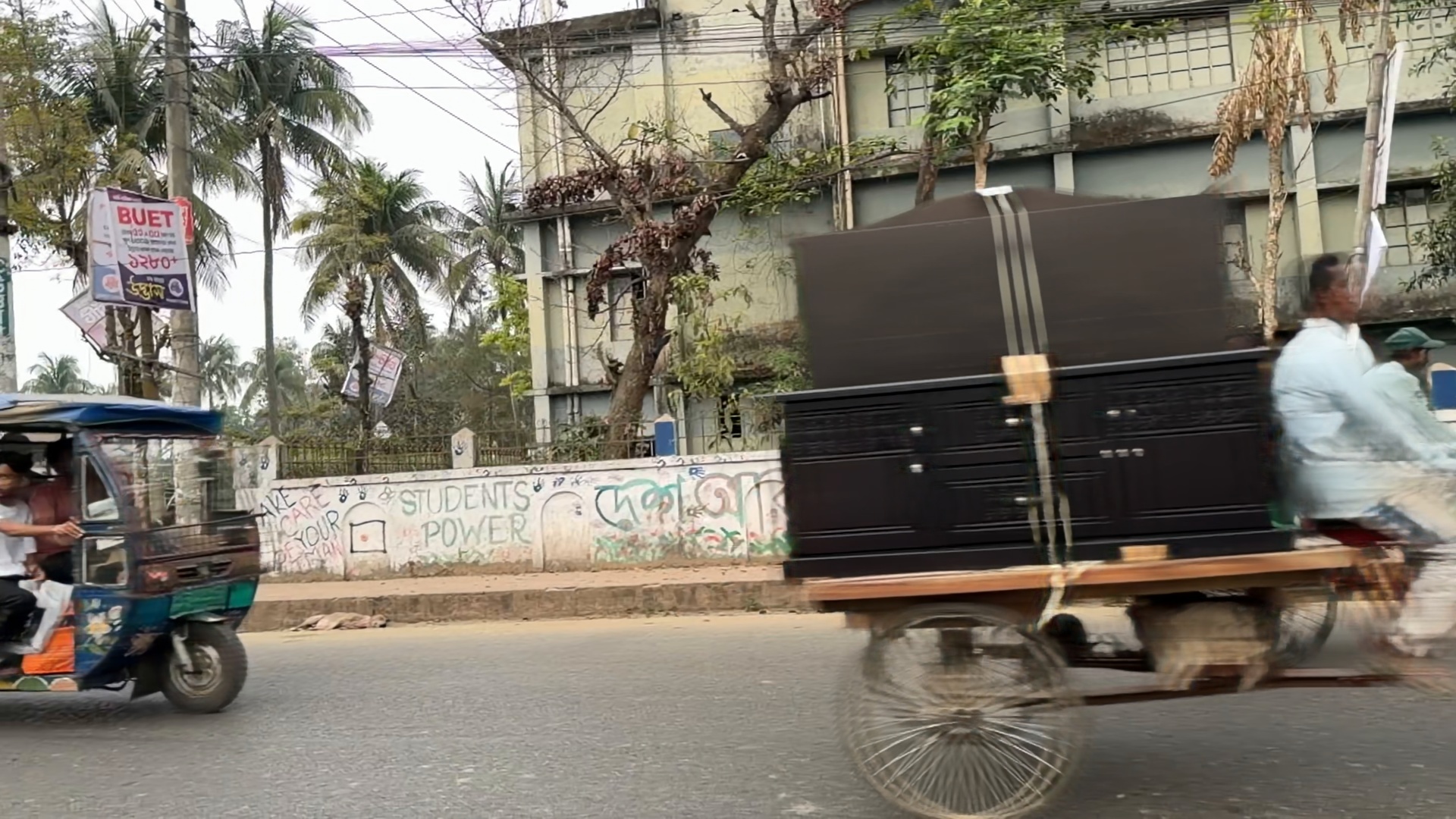}} &
\fcolorbox{mediumyellow}{white}{\includegraphics[width=0.192\textwidth]{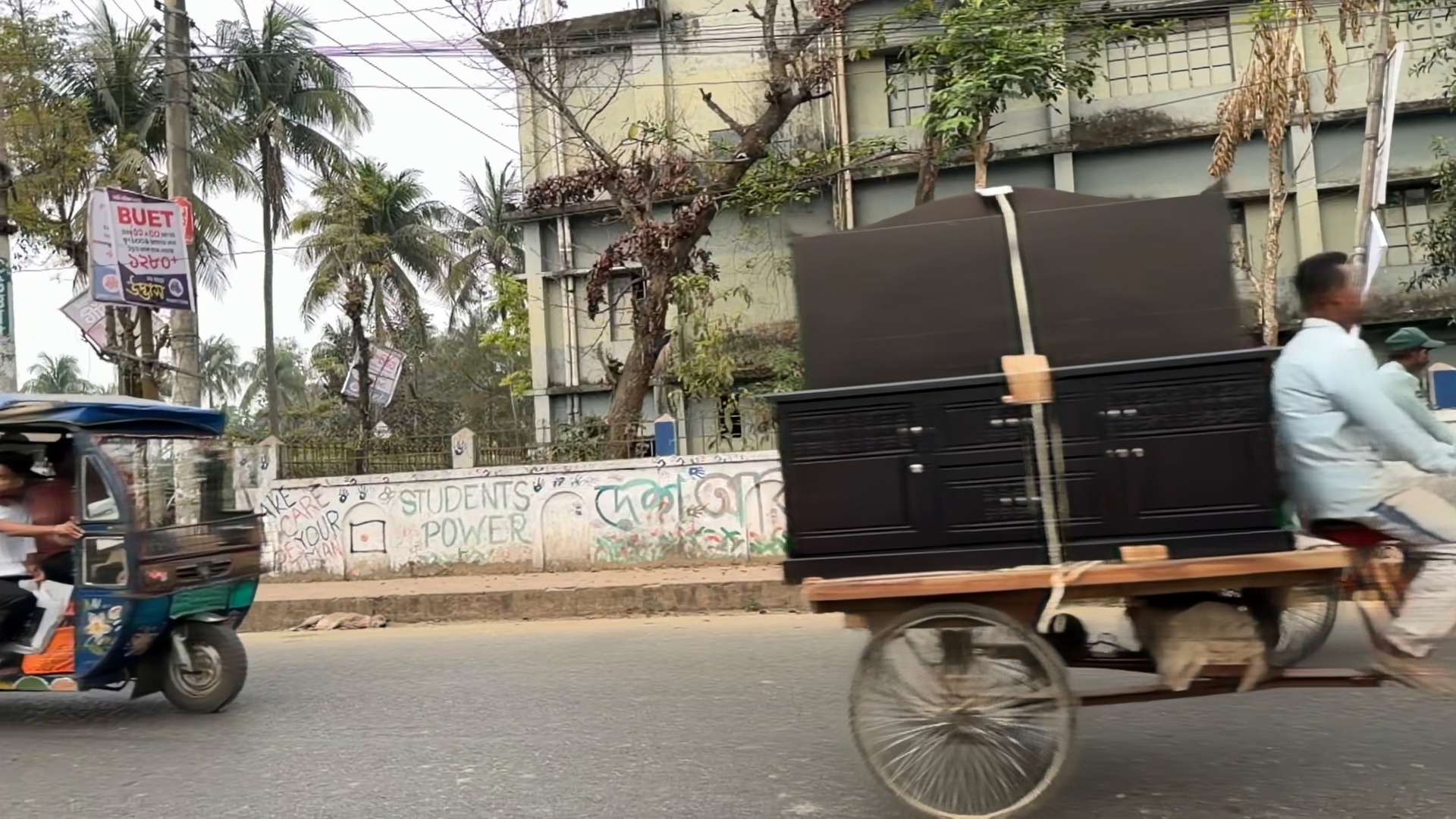}} &
\fcolorbox{mediumyellow}{white}{\includegraphics[width=0.192\textwidth]{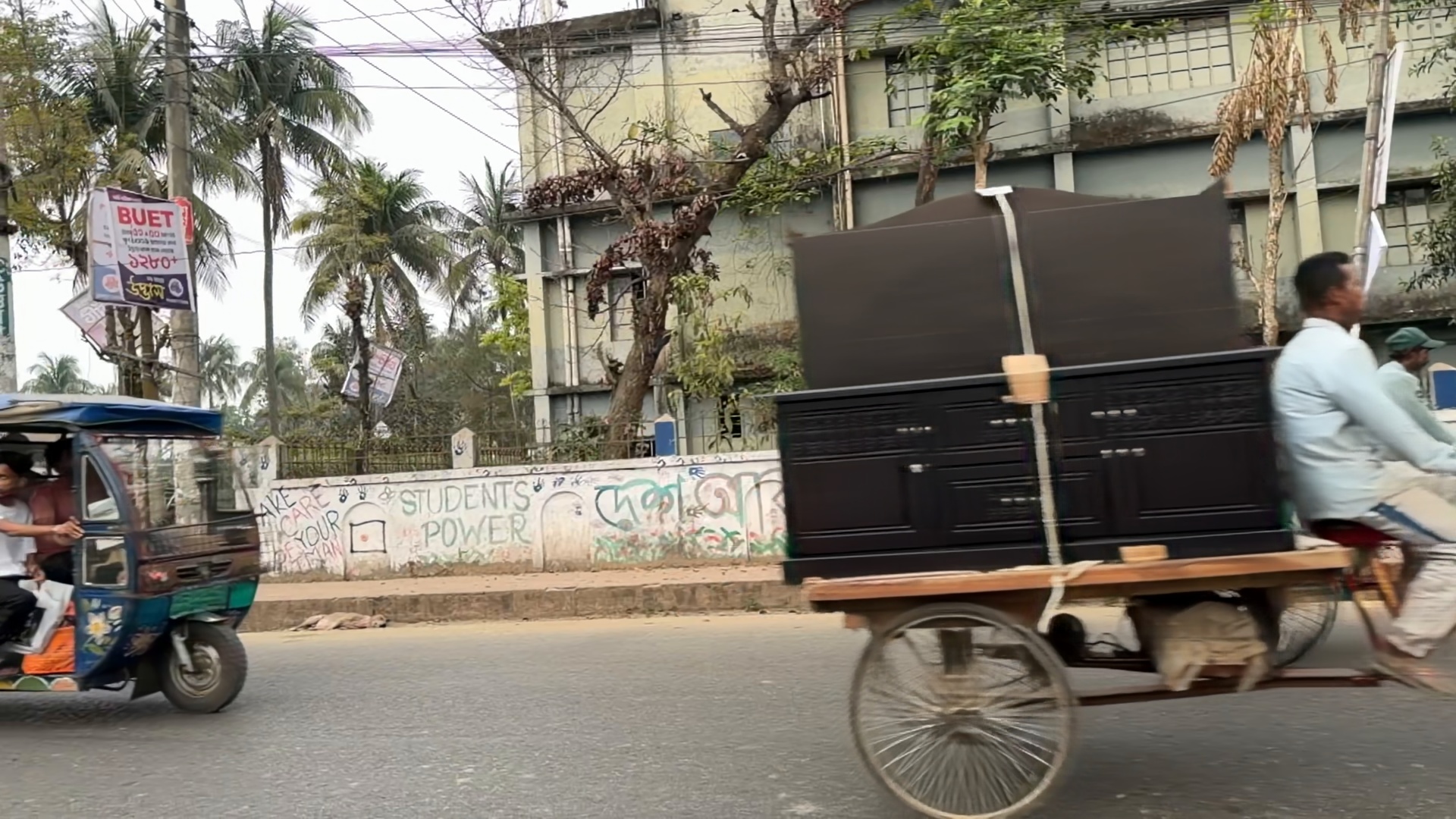}} &
\fcolorbox{mediumyellow}{white}{\includegraphics[width=0.192\textwidth]{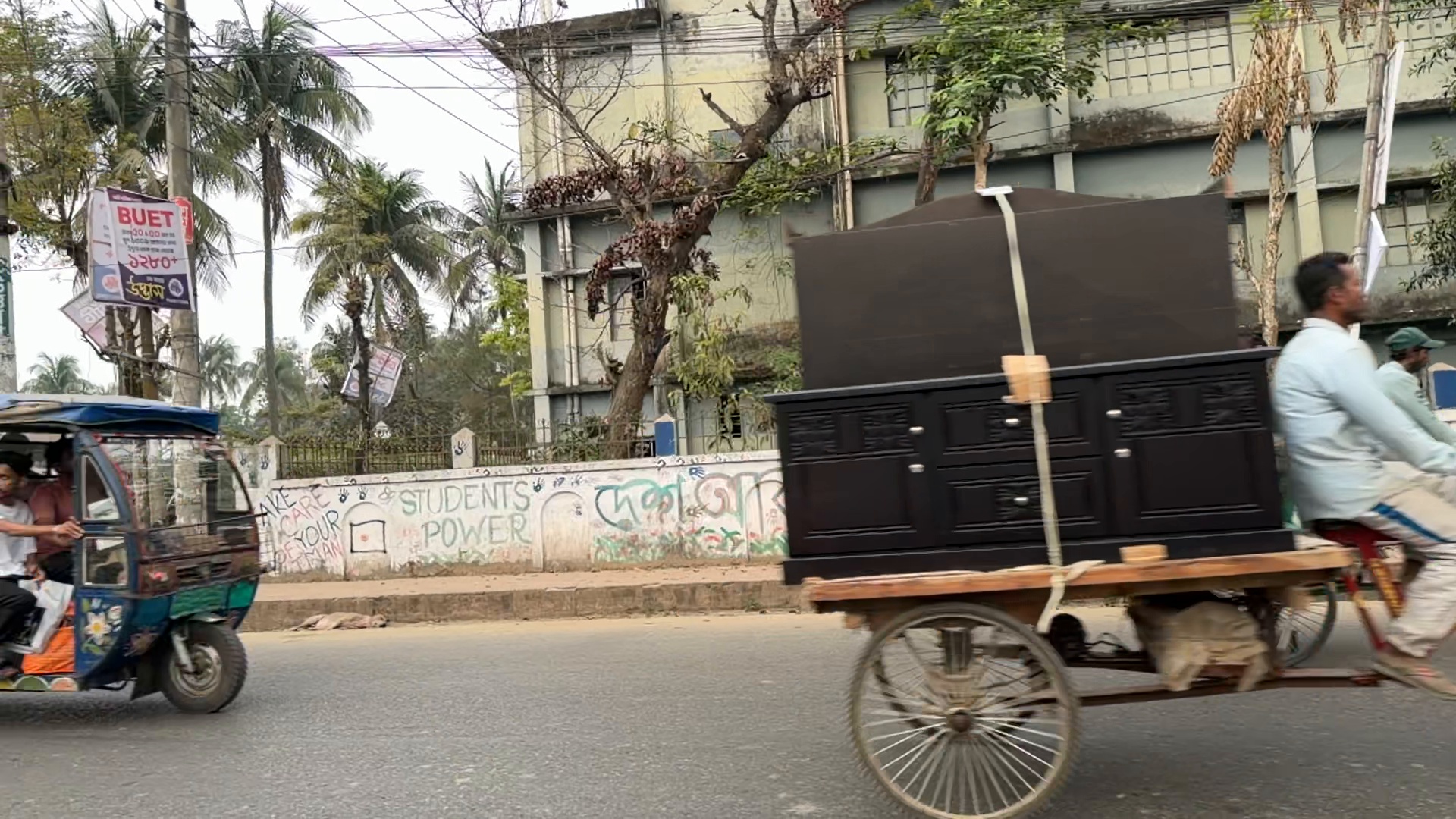}} \\[1.5mm]

\fcolorbox{hardred}{white}{\includegraphics[width=0.192\textwidth]{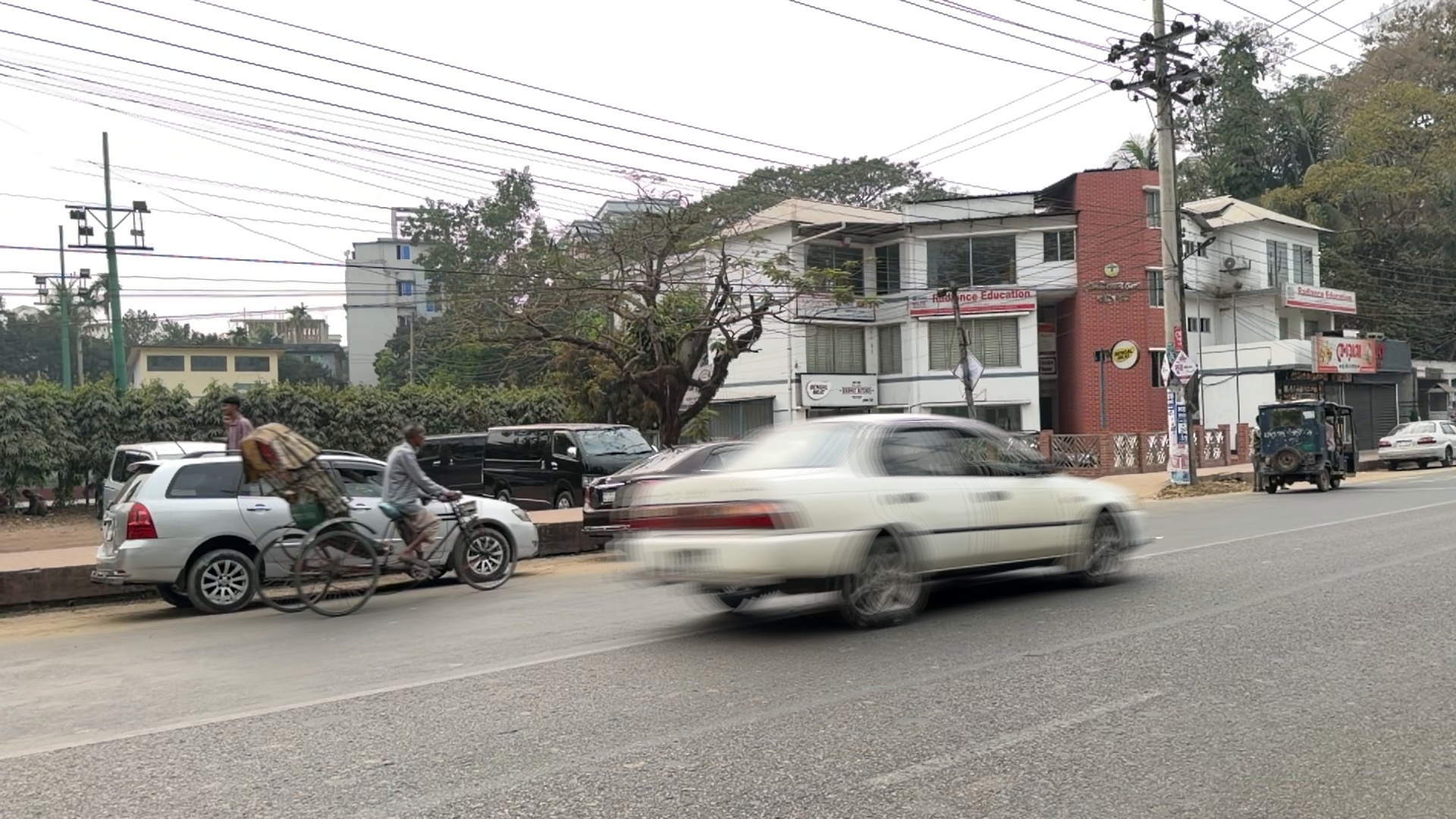}} &
\fcolorbox{hardred}{white}{\includegraphics[width=0.192\textwidth]{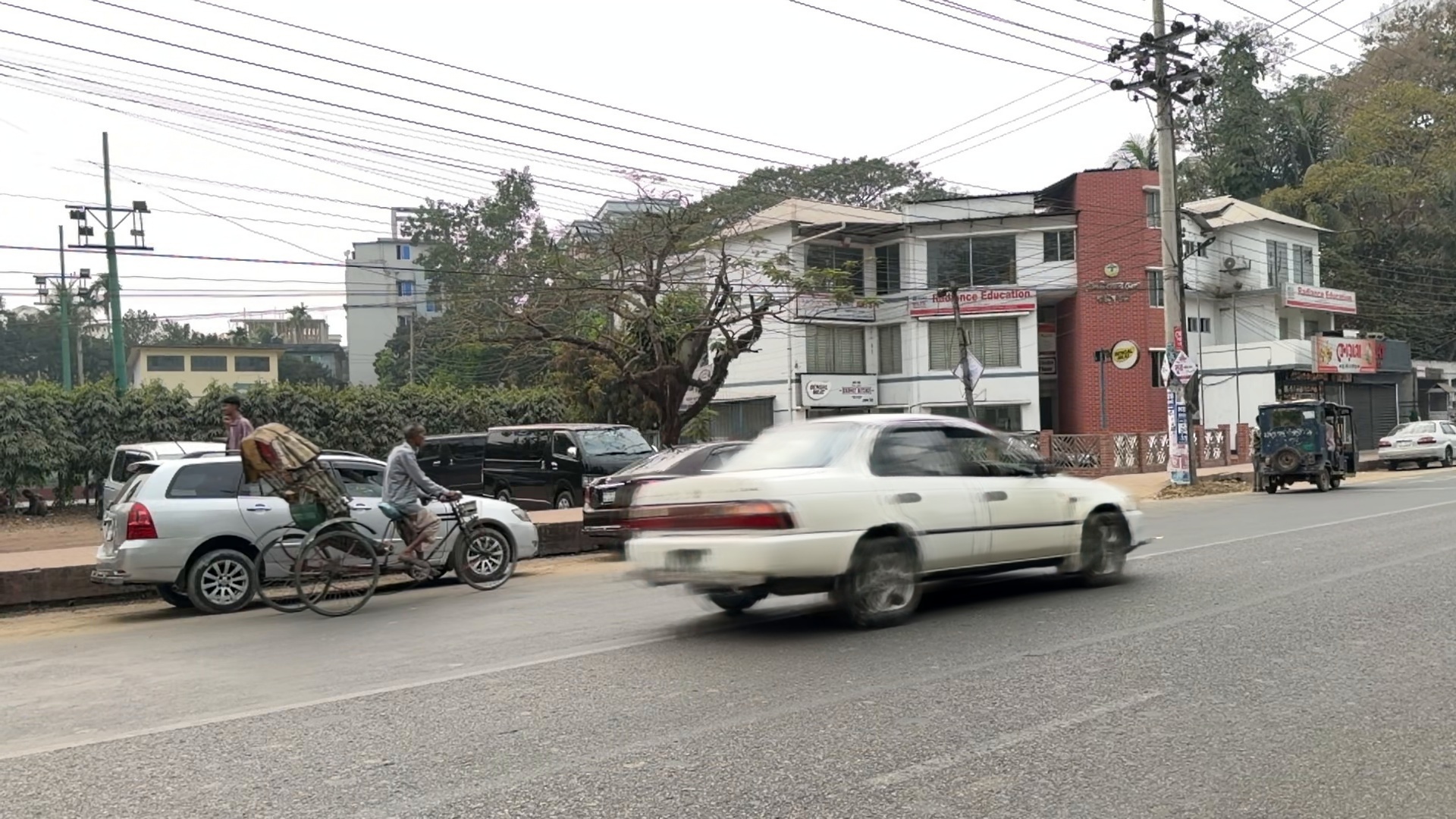}} &
\fcolorbox{hardred}{white}{\includegraphics[width=0.192\textwidth]{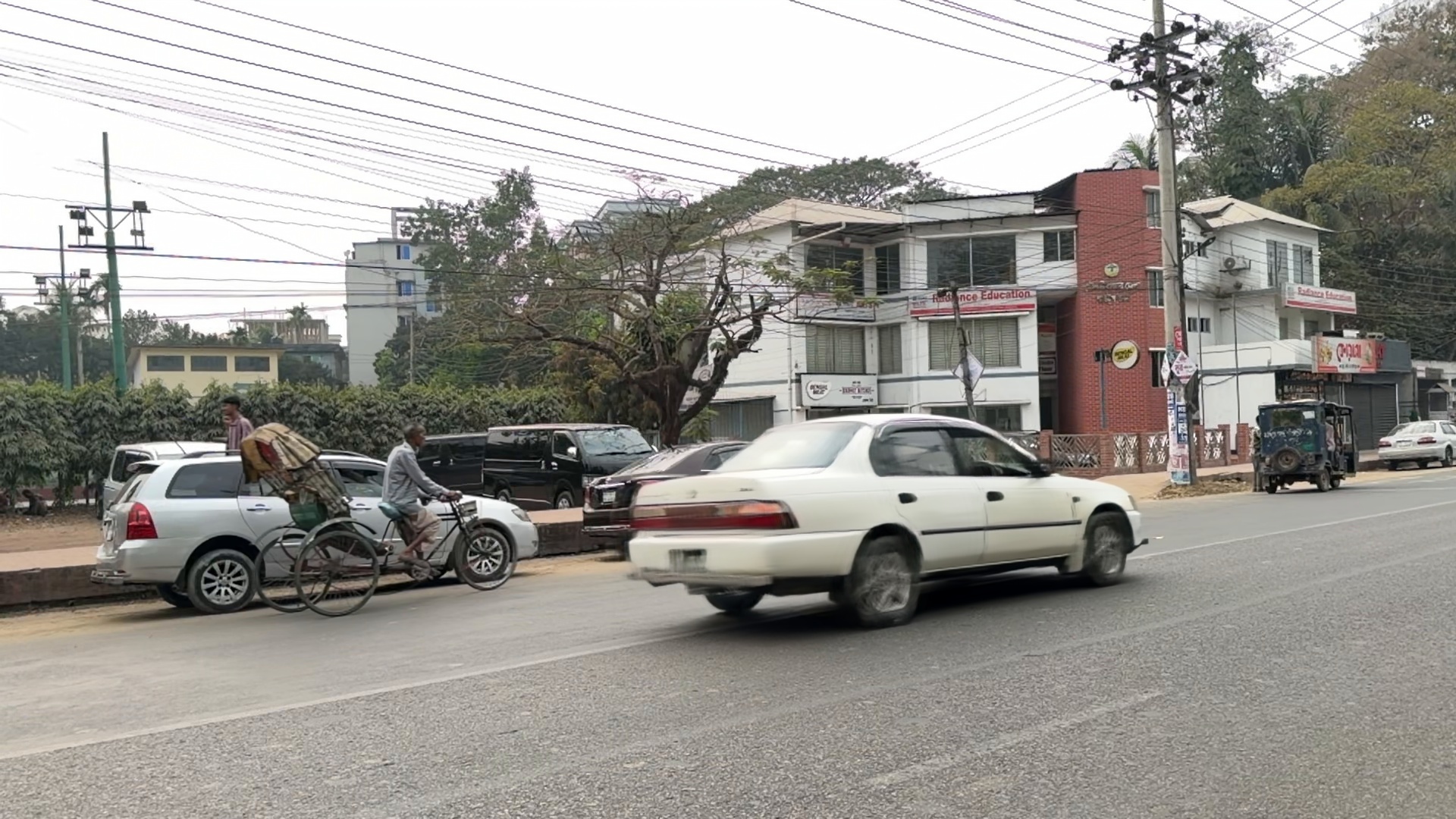}} &
\fcolorbox{hardred}{white}{\includegraphics[width=0.192\textwidth]{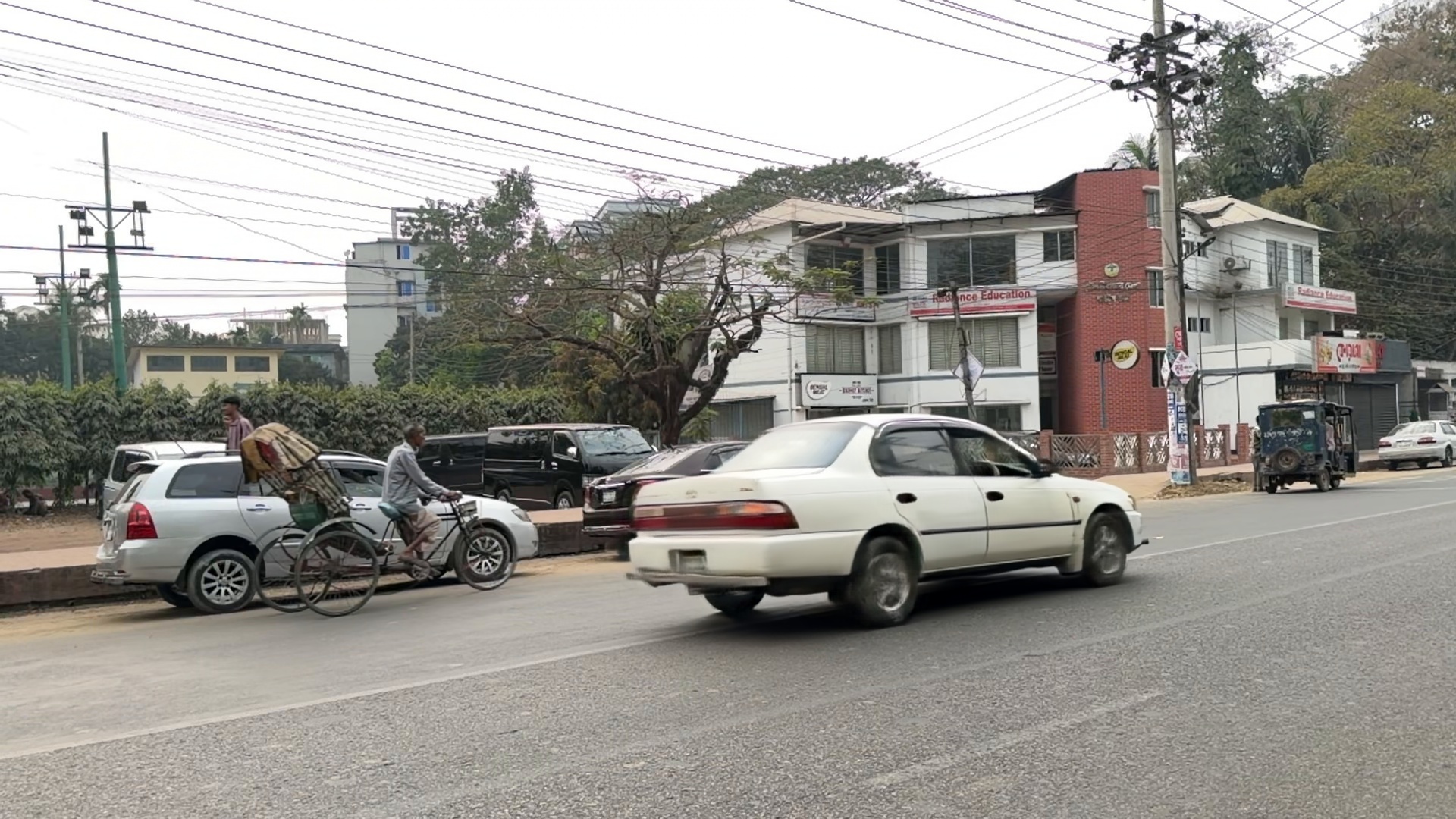}} &
\fcolorbox{hardred}{white}{\includegraphics[width=0.192\textwidth]{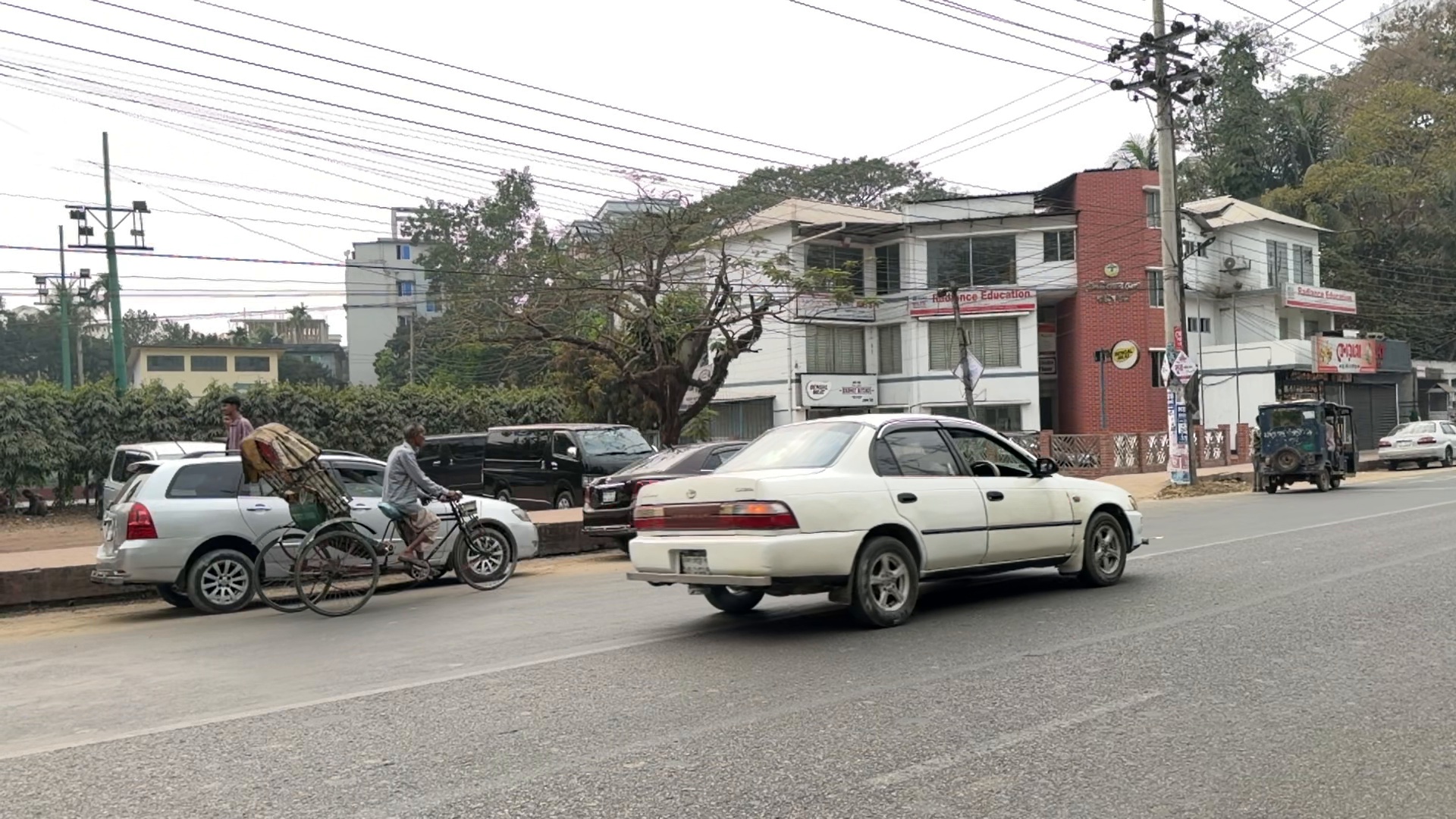}} \\[1.5mm]

\fcolorbox{hardred}{white}{\includegraphics[width=0.192\textwidth]{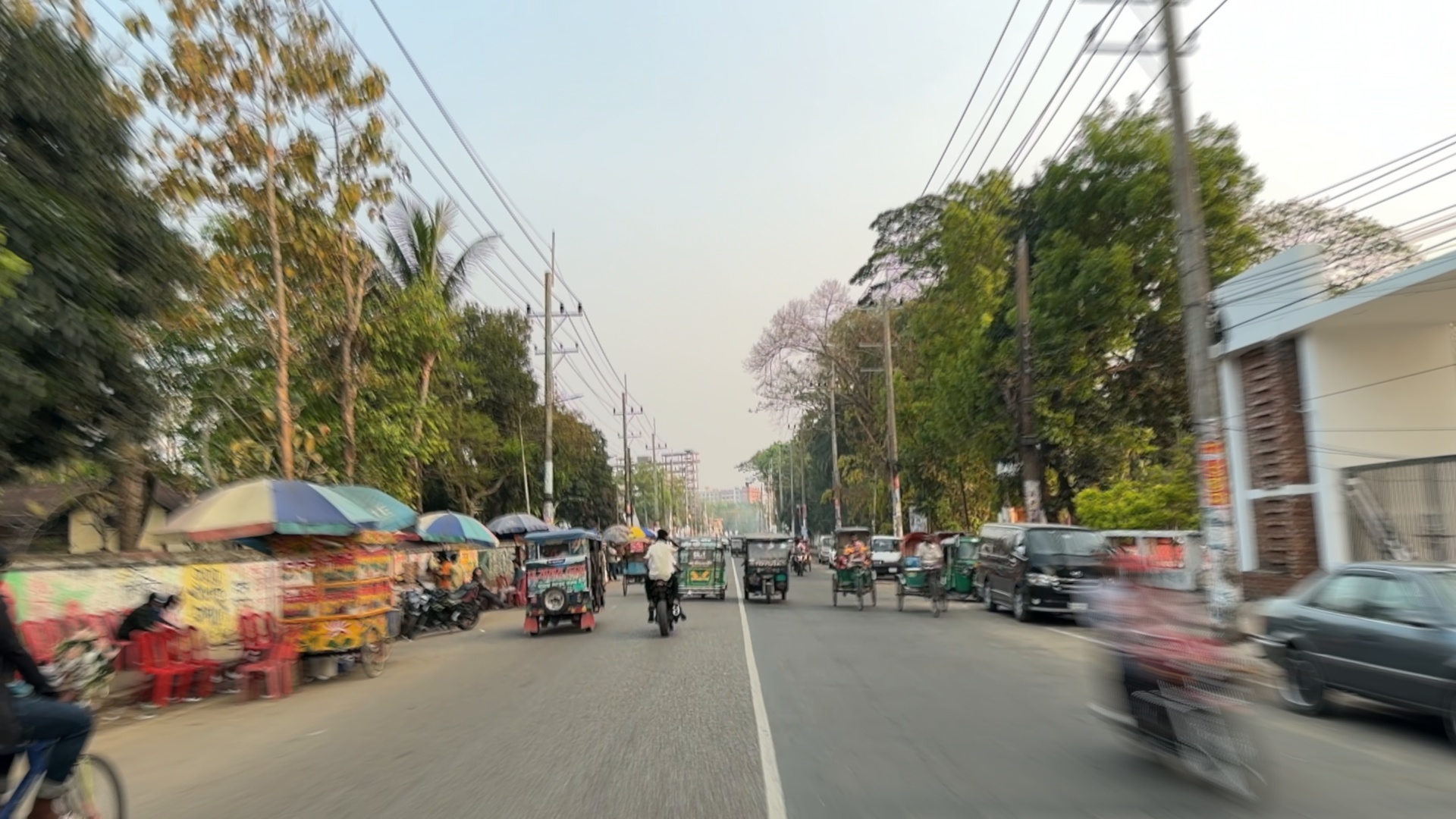}} &
\fcolorbox{hardred}{white}{\includegraphics[width=0.192\textwidth]{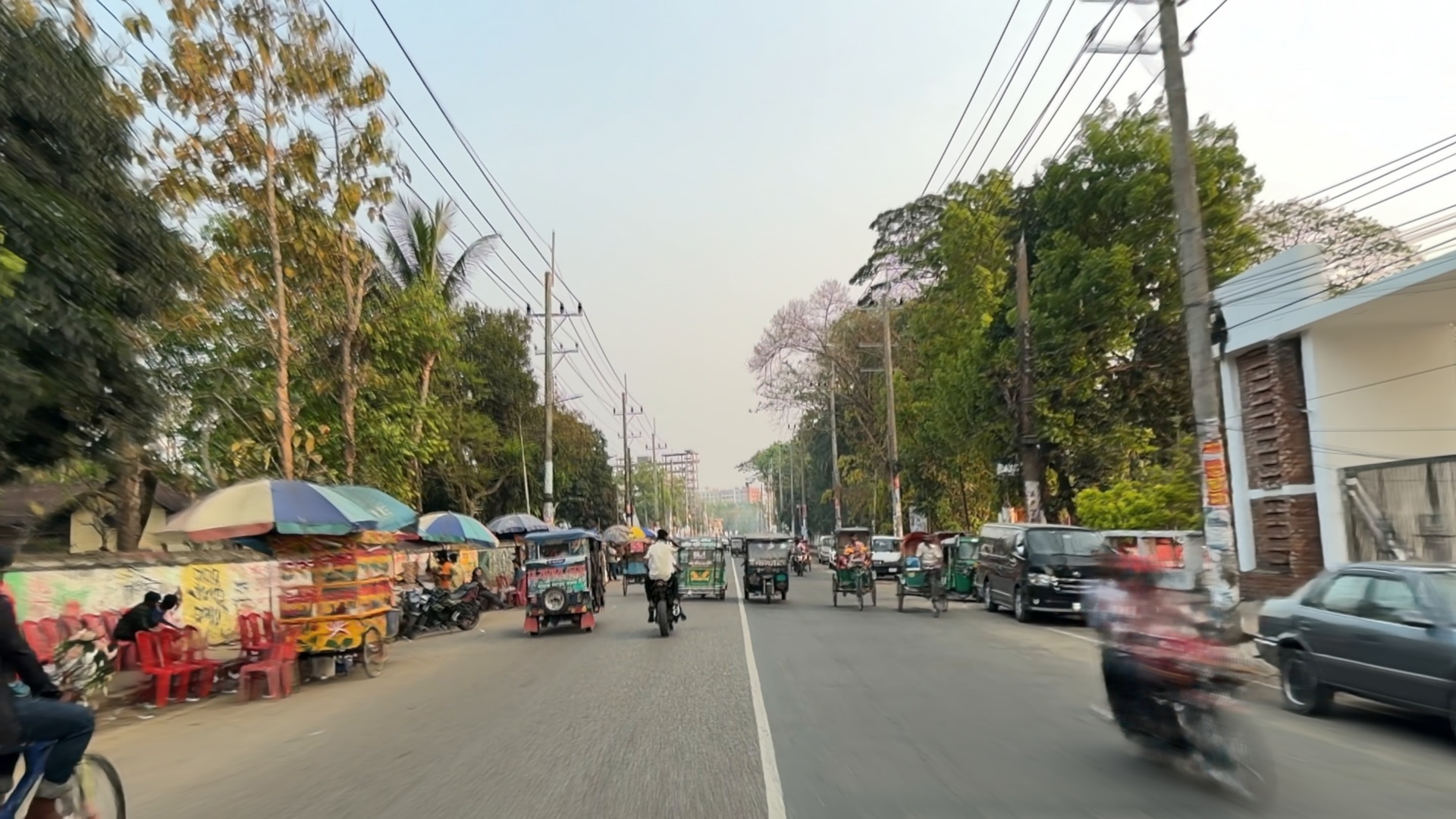}} &
\fcolorbox{hardred}{white}{\includegraphics[width=0.192\textwidth]{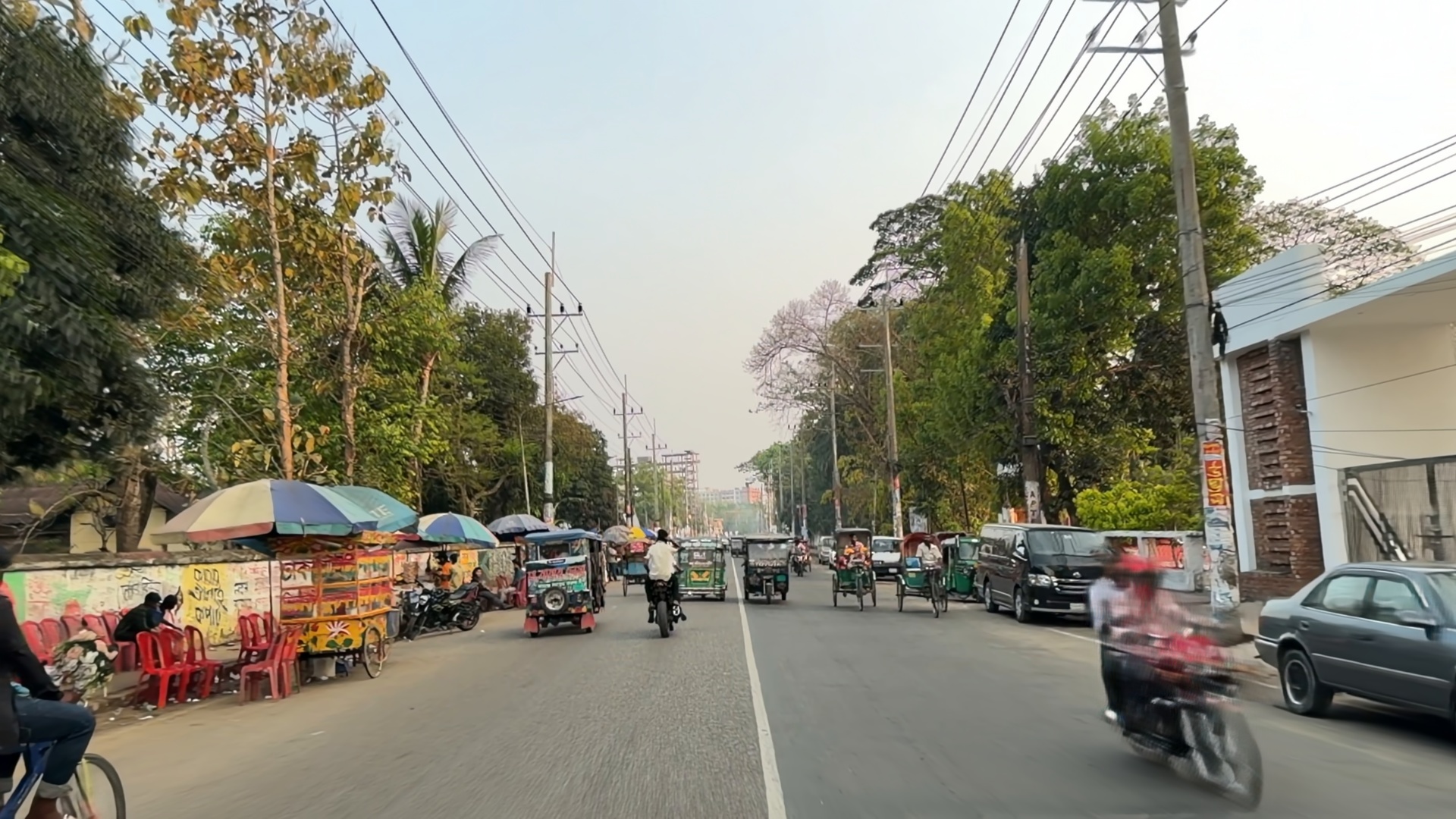}} &
\fcolorbox{hardred}{white}{\includegraphics[width=0.192\textwidth]{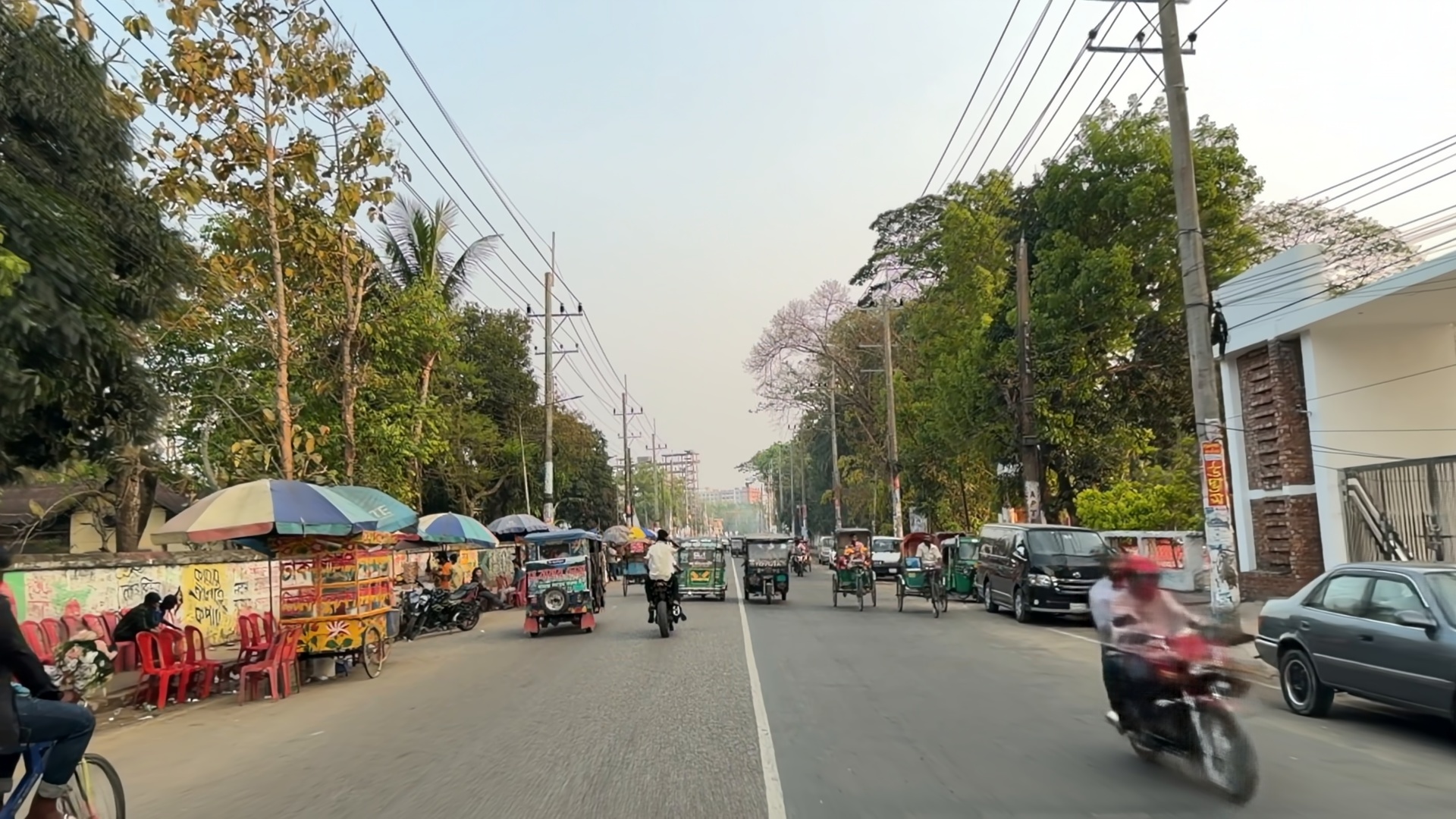}} &
\fcolorbox{hardred}{white}{\includegraphics[width=0.192\textwidth]{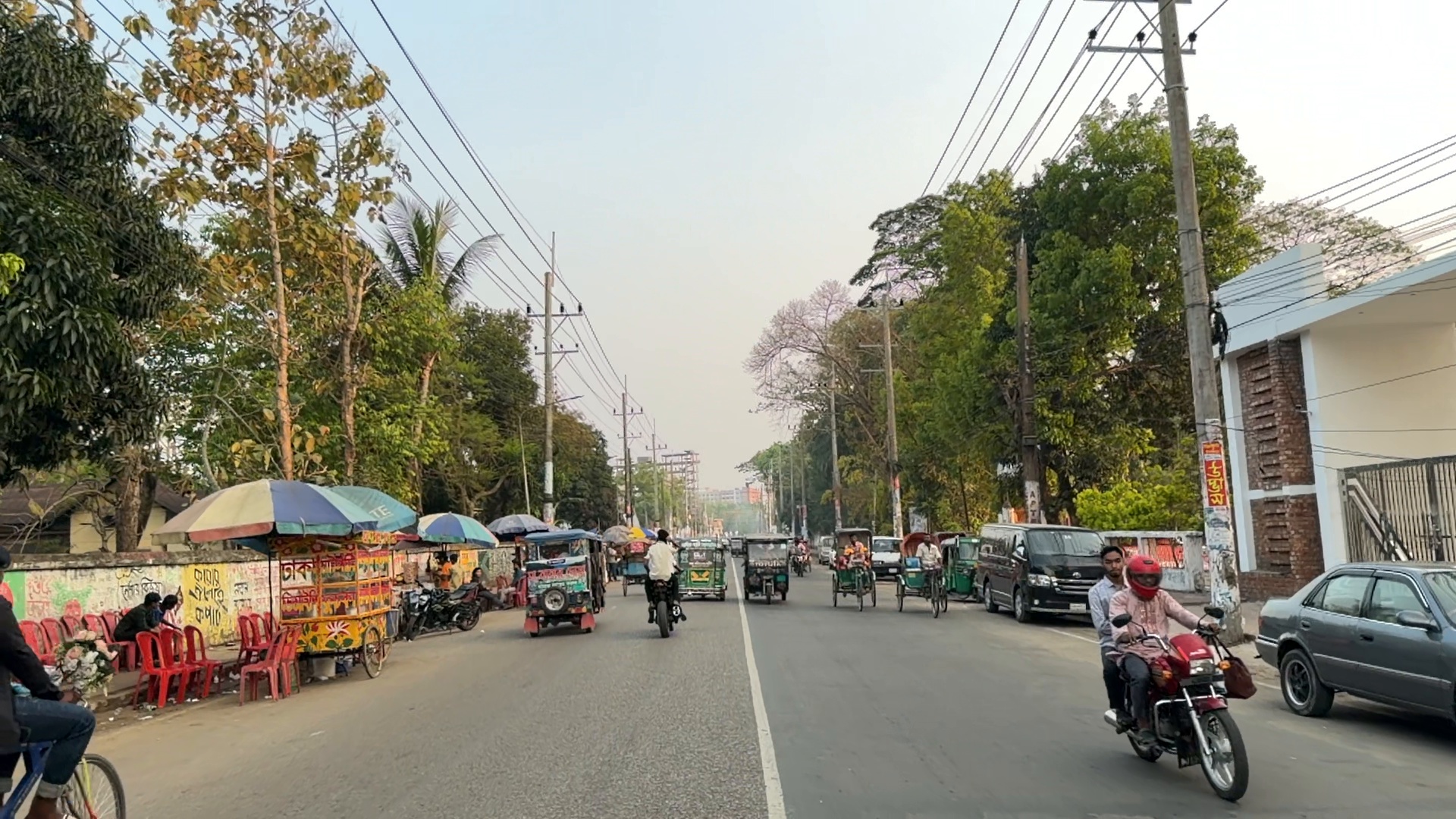}} \\

\end{tabular}

\caption{Qualitative comparison across difficulty levels. Easy sample (row 1, green,
6.0~px/frame): near-perfect restoration, PSNR 33.9--35.1~dB. Medium samples (rows 2--3,
yellow, 11.3~px/frame): emerging artifacts, PSNR 29.7--31.6~dB. Hard samples (rows 4--5,
red, 13.2~px/frame): severe residual blur with motion-dependent streaks,
PSNR 25.4--27.9~dB.}
\label{fig:qualitative}
\end{figure*}

Figure~\ref{fig:qualitative} visualizes stratified restoration quality. Easy samples
(row 1, 6.0~px/frame) restore cleanly with PSNR 33.9--35.1~dB; Medium samples
(rows 2--3, 11.3~px/frame) exhibit emerging artifacts with PSNR 29.7--31.6~dB; Hard
samples (rows 4--5, 13.2~px/frame) suffer severe residual blur and motion streaks with
PSNR 25.4--27.9~dB. This progressive degradation—clean restoration transitioning through
artifacts to fundamental failure—validates stratification captures perceptual breakpoints
correlating with motion magnitude.

\section{Discussion}

iPhoneBlur's stratified design enables deployment‑critical analysis missing from
aggregate‑only benchmarks. The 7.2–9.1 dB Easy‑to‑Hard degradation exceeding the
6.1–7.1 dB cross‑domain gap shows motion severity matters more than domain shift—
a finding entirely hidden by aggregates. Tiers correlate with optical flow
(6.0–13.2 px/frame) and adaptive window length (3–21 frames), enabling
confidence‑based routing: Easy samples on‑device, Hard offloaded to the cloud.
Per‑sample metadata supports ISP‑aware restoration jointly modeling blur and
sharpening, difficulty prediction, and progressive curricula that gradually
introduce harder samples. Cross‑domain experiments demonstrate bridging:
GoPro‑pretrained NAFNet drops 7.1 dB (33.7 → 26.6 dB) on iPhoneBlur, but
fine‑tuning recovers to 31.2 dB, cutting the gap to 2.5 dB and establishing an
effective transfer pathway between professional and consumer domains. By linking
performance to measurable scene properties, iPhoneBlur enables targeted diagnostics
for model weaknesses.

\section{Limitations}

Blur synthesis employs physically validated high-frequency suppression achieving
Cohen's $d=2.32$, statistically close to RealBlur‑J authentic blur
($d=2.24$, difference $<4\%$). The dataset focuses on iPhone 17 Pro to isolate
motion-based difficulty from confounding device variables. While the stratification
methodology relies on optical flow—a device‑independent motion cue—the dataset’s
consumer‑device characteristics (Apple ISP, iPhone 17 Pro) may differ from other
smartphone cameras; cross‑device validation remains future work. Content captures
typical smartphone photography scenarios (outdoor urban/suburban environments,
handheld motion). The adaptive windowing principles are general, but low‑light,
stabilized capture, and aerial footage remain untested and need validation.
Dataset scope prioritizes rigorous validation of stratification methodology over
exhaustive scenario coverage, establishing a foundation for community‑driven
extensions.

\section{Conclusion}

This work introduced iPhoneBlur, the first difficulty‑stratified consumer‑device
motion deblurring benchmark with rich per‑sample metadata. It contains 7,400
pairs from 51 iPhone 17 Pro videos (240 fps), partitioned into validated Easy
(17.3\%), Medium (52.6\%), and Hard (30.1\%) levels. Stratification is confirmed by
optical flow correlation ($\rho=-0.41$, $p<10^{-294}$), a $2.2\times$ monotonic
motion increase, human perceptual agreement ($\rho=0.74$, $p<0.001$), and realistic
high‑frequency suppression (Cohen's $d=2.32$ vs.\ RealBlur‑J $d=2.24$). Baseline
evaluation reveals a consistent 7.2–9.1 dB Easy‑to‑Hard degradation—far exceeding
aggregate variation—validating that stratification exposes fundamental restoration
challenges hidden by standard reporting. Beyond benchmarking, iPhoneBlur enables
difficulty‑aware adaptive inference, ISP‑aware restoration, progressive training
curricula, and professional‑to‑consumer domain adaptation.

\section*{Ethics Statement}

All videos were recorded by the authors in public outdoor environments (streets, parks,
plazas) in compliance with local regulations. No footage focuses on individuals, and
no personally identifiable information is present. The dataset contains no close‑up
footage, no indoor private spaces, and no personally identifiable information. Usage
terms explicitly prohibit surveillance targeting individuals, discriminatory algorithms,
unauthorised biometric identification, and privacy‑violating applications. This work
advances mobile computational photography for improved motion‑affected image quality.

\section*{Data and Code Availability}

\textbf{Dataset:} \url{https://kaggle.com/datasets/shafi09/iphoneblur} (CC BY 4.0). \\
\textbf{Code:} \url{https://github.com/C-loud-Nine/iPhoneBlur} (MIT).

\begin{ack}
This research received no external funding. The authors declare no competing interests.
\end{ack}

\bibliographystyle{unsrtnat}
\bibliography{references}

\clearpage
\appendix

\noindent{\Large\bfseries Appendix}

\section{Training Configuration and Reproducibility}
\label{app:training}

All methods fine‑tune from official GoPro‑pretrained weights on NVIDIA Tesla T4 GPUs
(16\,GB VRAM) using standardised configurations for fair comparison. Training employs
patch size $256\times256$, AdamW optimizer ($\beta_1=0.9$, $\beta_2=0.999$, weight decay
$10^{-4}$), cosine annealing learning rate schedule with $\eta_{\min}=10^{-6}$, and
standard data augmentation (random horizontal flips, vertical flips).
Training data comprises 5,714 image pairs from 39 videos; the test set contains 1,686
pairs from 12 videos.

\subsection{Architecture‑Specific Configurations}

The number of parameter updates was kept roughly comparable across architectures
($2.5\times10^4$–$1.7\times10^5$ updates). All models share the same initial learning rate
$1\times10^{-4}$, AdamW optimizer ($\beta_1=0.9$, $\beta_2=0.999$, weight decay $10^{-4}$),
cosine annealing schedule ($\eta_{\min}=10^{-6}$), and the $\mathcal{L}_1$ loss (MIMO‑UNet
uses multi‑scale $\mathcal{L}_1$). Restormer and FFTformer additionally used mixed‑precision
training (FP16). Epoch counts were chosen based on preliminary convergence runs; specific
values are listed below.

\noindent\textbf{CNN‑based architectures:}
\begin{itemize}[noitemsep,topsep=2pt]
\item \textbf{NAFNet}~\cite{chen2022simple} (67.9M parameters, width=64): effective batch size 8 (2 physical $\times$ 4 accumulation), 60 epochs.
\item \textbf{HINet}~\cite{chen2021hinet} (26.5M): batch size 4, 60 epochs.
\item \textbf{MIMO‑UNet}~\cite{cho2021rethinking} (16.8M): batch size 4, 60 epochs.
\end{itemize}

\noindent\textbf{Transformer‑based architectures:}
\begin{itemize}[noitemsep,topsep=2pt]
\item \textbf{Restormer}~\cite{zamir2022restormer} (26.1M): effective batch size 4 (2 physical $\times$ 2 accumulation), 30 epochs, gradient clipping (max norm 1.0).
\item \textbf{FFTformer}~\cite{kong2022efficient} (24.7M): effective batch size 4 (2 physical $\times$ 2 accumulation), 30 epochs.
\item \textbf{Instruct‑IR}~\cite{conde2024instructir} (22.3M): batch size 2, 50 epochs, prompt ``Restore the blurred image''.
\end{itemize}

Gradient accumulation was used where needed to fit larger effective batch sizes within
GPU memory limits. For each architecture, the reported epoch count corresponds to the
point where validation PSNR stopped improving meaningfully; further training yielded
negligible gains ($<0.1$ dB). All models were trained to convergence under the stated
settings.

All experiments run with PyTorch 1.13+ and CUDA 11.7+. Official implementations are used
without architectural modifications. During training, patches are randomly cropped; at
evaluation time, full‑resolution images are processed without cropping. Inputs are
normalised to $[0,1]$.

\subsection{Multi‑Run Reproducibility}

To assess training stability, each method is independently fine‑tuned from scratch three
times with different random seeds (42, 123, 456). The sample standard deviation across
the three runs is reported alongside the mean for every metric and difficulty level,
using the same full‑resolution test set as in the main paper (Section~5.3). This approach
follows standard practice in image restoration and provides a direct estimate of the
variability introduced by weight initialisation and data ordering.

Tables~\ref{tab:psnr_multirun}--\ref{tab:lpips_multirun} present the complete
difficulty‑stratified results with $\pm 1\sigma$ uncertainty. Observed standard
deviations of 0.2–0.4 dB (PSNR), 0.004–0.012 (SSIM), and 0.002–0.005 (LPIPS) are
typical for deep‑learning‑based deblurring and confirm that the conclusions drawn in the
main paper are robust to initialisation noise.

\begin{table}[h]
\centering
\caption{Multi‑run PSNR (dB) statistics. Mean $\pm$ standard deviation computed from
three independent training runs evaluated on the full‑resolution test set.}
\label{tab:psnr_multirun}
\small
\begin{tabular}{lcccc}
\toprule
Method & Overall & Easy & Medium & Hard \\
\midrule
NAFNet      & $31.2 \pm 0.2$ & $35.1 \pm 0.3$ & $31.6 \pm 0.2$ & $27.9 \pm 0.3$ \\
HINet       & $30.5 \pm 0.2$ & $34.6 \pm 0.3$ & $30.9 \pm 0.2$ & $27.1 \pm 0.3$ \\
Restormer   & $30.9 \pm 0.2$ & $34.9 \pm 0.3$ & $31.2 \pm 0.2$ & $27.5 \pm 0.3$ \\
MIMO-UNet   & $29.3 \pm 0.3$ & $33.9 \pm 0.2$ & $29.7 \pm 0.3$ & $25.7 \pm 0.4$ \\
Instruct-IR & $29.5 \pm 0.3$ & $34.5 \pm 0.2$ & $29.9 \pm 0.3$ & $25.4 \pm 0.4$ \\
FFTformer   & $31.0 \pm 0.2$ & $35.0 \pm 0.3$ & $31.3 \pm 0.2$ & $27.7 \pm 0.3$ \\
\bottomrule
\end{tabular}
\end{table}

\begin{table}[h]
\centering
\caption{Multi‑run SSIM statistics. Standard deviations of 0.004–0.012 indicate highly
consistent structural preservation across runs.}
\label{tab:ssim_multirun}
\small
\begin{tabular}{lcccc}
\toprule
Method & Overall & Easy & Medium & Hard \\
\midrule
NAFNet      & $0.945 \pm 0.004$ & $0.978 \pm 0.002$ & $0.953 \pm 0.004$ & $0.907 \pm 0.008$ \\
HINet       & $0.938 \pm 0.005$ & $0.975 \pm 0.003$ & $0.947 \pm 0.005$ & $0.895 \pm 0.009$ \\
Restormer   & $0.940 \pm 0.004$ & $0.977 \pm 0.002$ & $0.949 \pm 0.004$ & $0.899 \pm 0.008$ \\
MIMO-UNet   & $0.920 \pm 0.007$ & $0.973 \pm 0.003$ & $0.933 \pm 0.006$ & $0.861 \pm 0.012$ \\
Instruct-IR & $0.924 \pm 0.006$ & $0.978 \pm 0.003$ & $0.935 \pm 0.006$ & $0.869 \pm 0.011$ \\
FFTformer   & $0.943 \pm 0.004$ & $0.977 \pm 0.002$ & $0.951 \pm 0.004$ & $0.903 \pm 0.008$ \\
\bottomrule
\end{tabular}
\end{table}

\begin{table}[h]
\centering
\caption{Multi‑run LPIPS (lower is better) statistics. Standard deviations of 0.002–0.005
confirm stable perceptual quality across training runs.}
\label{tab:lpips_multirun}
\small
\begin{tabular}{lcccc}
\toprule
Method & Overall & Easy & Medium & Hard \\
\midrule
NAFNet      & $0.049 \pm 0.002$ & $0.020 \pm 0.001$ & $0.042 \pm 0.002$ & $0.080 \pm 0.004$ \\
HINet       & $0.056 \pm 0.003$ & $0.023 \pm 0.002$ & $0.048 \pm 0.003$ & $0.092 \pm 0.005$ \\
Restormer   & $0.053 \pm 0.002$ & $0.022 \pm 0.001$ & $0.046 \pm 0.002$ & $0.088 \pm 0.004$ \\
MIMO-UNet   & $0.075 \pm 0.005$ & $0.026 \pm 0.002$ & $0.065 \pm 0.004$ & $0.127 \pm 0.007$ \\
Instruct-IR & $0.071 \pm 0.004$ & $0.019 \pm 0.002$ & $0.063 \pm 0.004$ & $0.118 \pm 0.006$ \\
FFTformer   & $0.050 \pm 0.002$ & $0.021 \pm 0.001$ & $0.043 \pm 0.002$ & $0.083 \pm 0.004$ \\
\bottomrule
\end{tabular}
\end{table}

\noindent The small standard deviations across all metrics and difficulty levels
demonstrate that the fine‑tuning procedure is stable and that the observed performance
trends (substantial Easy‑to‑Hard degradation, architecture‑agnostic behaviour) are not
artefacts of particular random initializations.

\subsection{Baseline Implementation Details}

All models are initialised from official GoPro‑pretrained checkpoints and evaluated on
full‑resolution images without cropping. Official repositories and licenses:
NAFNet (\url{https://github.com/megvii-research/NAFNet}, Apache~2.0),
HINet (\url{https://github.com/megvii-model/HINet}, MIT),
Restormer (\url{https://github.com/swz30/Restormer}, MIT),
MIMO‑UNet (\url{https://github.com/chosj95/MIMO-UNet}, Unlicensed),
Instruct‑IR (\url{https://github.com/mv-lab/InstructIR}, MIT),
FFTformer (\url{https://github.com/kkkls/FFTformer}, MIT).

\section{Synthesis Quality Validation}
\label{app:cohens_d}

High‑frequency suppression analysis validates synthesis realism through physics‑based
temporal integration metrics. High‑frequency energy is computed as the mean absolute
Laplacian of the luminance channel (ITU‑R BT.601). Cohen's $d$ effect size quantifies
suppression strength by comparing this energy between sharp and blurred images:
\begin{equation}
d = \frac{\mu_{\text{sharp}} - \mu_{\text{blur}}}{\sigma_{\text{pooled}}}
\end{equation}
where $\mu$ denotes mean high‑frequency energy and $\sigma_{\text{pooled}}$ is the
pooled standard deviation.

Table~\ref{tab:cohens_appendix} shows Cohen's $d$ increasing monotonically across
difficulty levels, confirming the adaptive window methodology produces physically
consistent blur. Train‑test consistency validates generalisation.

\begin{table}[h]
\centering
\caption{Cohen's $d$ effect size for high‑frequency suppression by difficulty level and
data split. Monotonic increase validates difficulty stratification reflects physical
motion characteristics. Overall $d=2.32$ approaches RealBlur‑J authentic blur ($d=2.24$).}
\label{tab:cohens_appendix}
\small
\begin{tabular}{lcccc}
\toprule
Split & Easy & Medium & Hard & Overall \\
\midrule
Train    & 2.10 & 2.50 & 2.64 & 2.34 \\
Test     & 1.84 & 2.23 & 2.63 & 2.27 \\
\midrule
Combined & 2.02 & 2.45 & 2.64 & \textbf{2.32} \\
\bottomrule
\end{tabular}
\end{table}

\textbf{Cross‑benchmark comparison:} iPhoneBlur synthesis quality ($d=2.32$) matches
established benchmarks: GoPro $d=2.81$, HIDE $d=2.79$, REDS $d=2.45$, RealBlur‑J
$d=2.24$. RealBlur‑R exhibits weaker suppression ($d=0.68$) due to less severe blur.

\section{Dataset Diversity and Content Analysis}
\label{app:diversity}

Table~\ref{tab:video_diversity} quantifies source video contribution and cross‑difficulty
coverage. The critical finding is that 78\% of source videos (40/51) contribute samples
to all three difficulty tiers, preventing spurious content‑difficulty correlation.

\begin{table}[h]
\centering
\caption{Source video diversity and cross‑difficulty coverage. High cross‑tier contribution
(78\% of videos span all difficulties) validates motion‑based stratification rather than
content‑based clustering.}
\label{tab:video_diversity}
\small
\begin{tabular}{lcccc}
\toprule
Metric & Total & Easy & Medium & Hard \\
\midrule
Source videos & 51 & 41 & 50 & 50 \\
Total samples & 7,400 & 1,277 & 3,895 & 2,228 \\
\midrule
Videos spanning all 3 tiers & 40 (78\%) & \multicolumn{3}{c}{---} \\
Videos spanning 2 tiers   & 10 (20\%) & \multicolumn{3}{c}{---} \\
Videos spanning 1 tier    &  1 (2\%)  & \multicolumn{3}{c}{---} \\
\bottomrule
\end{tabular}
\end{table}

\textbf{Content distribution:} The dataset captures diverse smartphone photography
scenarios including outdoor urban scenes, natural environments, human activities, and
dynamic motion. Content spans typical handheld capture conditions with varied lighting
(daylight, overcast, golden hour) and motion sources (camera shake, panning, subject
motion).

\textbf{Difficulty‑content independence:} All major scene categories exhibit similar
difficulty distributions (approximately 50--55\% Medium, 17--19\% Easy, 28--32\% Hard),
confirming stratification reflects motion dynamics rather than semantic content.

\section{Comprehensive Metadata Documentation}
\label{app:metadata}

Each sample includes metadata spanning quality metrics, physical scene properties, ISP
characteristics, and synthesis parameters. Table~\ref{tab:metadata_appendix} documents
the primary fields with observed ranges from the dataset.

\begin{table}[h]
\centering
\caption{Primary metadata fields per sample. Ranges reflect actual dataset statistics,
enabling ISP‑aware restoration and difficulty prediction research.}
\label{tab:metadata_appendix}
\small
\begin{tabular}{llp{5.5cm}}
\toprule
Category & Field & Description \& Range \\
\midrule
Identifiers & \texttt{img\_id} & Sample identifier (0--7399) \\
& \texttt{video} & Source video filename \\
& \texttt{img\_num} & Frame number in source video \\
\midrule
Quality & \texttt{psnr} & PSNR (dB): 20.1--35.0 \\
& \texttt{ssim} & SSIM: 0.65--0.98 \\
& \texttt{lpips} & LPIPS: 0.013--0.350 \\
\midrule
Physical & \texttt{motion} & Optical flow (px/frame): 0.0--28.1 \\
& \texttt{sharpness} & Laplacian variance \\
& \texttt{contrast} & RMS luminance \\
\midrule
ISP & \texttt{isp\_sharp} & High‑frequency energy in sharp image (mean abs.\ Laplacian) \\
& \texttt{isp\_blur} & High‑frequency energy in blur image (always lower than sharp) \\
& \texttt{isp\_diff} & Suppression magnitude (sharp $-$ blur) \\
\midrule
Synthesis & \texttt{blur\_window} & Window size: 3--21 frames \\
& \texttt{difficulty} & Easy / Medium / Hard \\
\bottomrule
\end{tabular}
\end{table}

\textbf{Statistics by difficulty:}
\begin{itemize}[noitemsep,topsep=2pt]
\item \textbf{Easy} (17.3\%, 1,277 samples): PSNR $31.7 \pm 1.2$~dB, SSIM $0.96 \pm 0.01$, optical flow $6.0 \pm 5.0$~px/frame
\item \textbf{Medium} (52.6\%, 3,895 samples): PSNR $26.5 \pm 1.6$~dB, SSIM $0.87 \pm 0.05$, optical flow $11.3 \pm 5.6$~px/frame
\item \textbf{Hard} (30.1\%, 2,228 samples): PSNR $22.6 \pm 0.9$~dB, SSIM $0.75 \pm 0.06$, optical flow $13.2 \pm 4.3$~px/frame
\end{itemize}

\section{Qualitative Analysis}
\label{app:qualitative}

\begin{figure}[t]
\centering
\includegraphics[width=\textwidth]{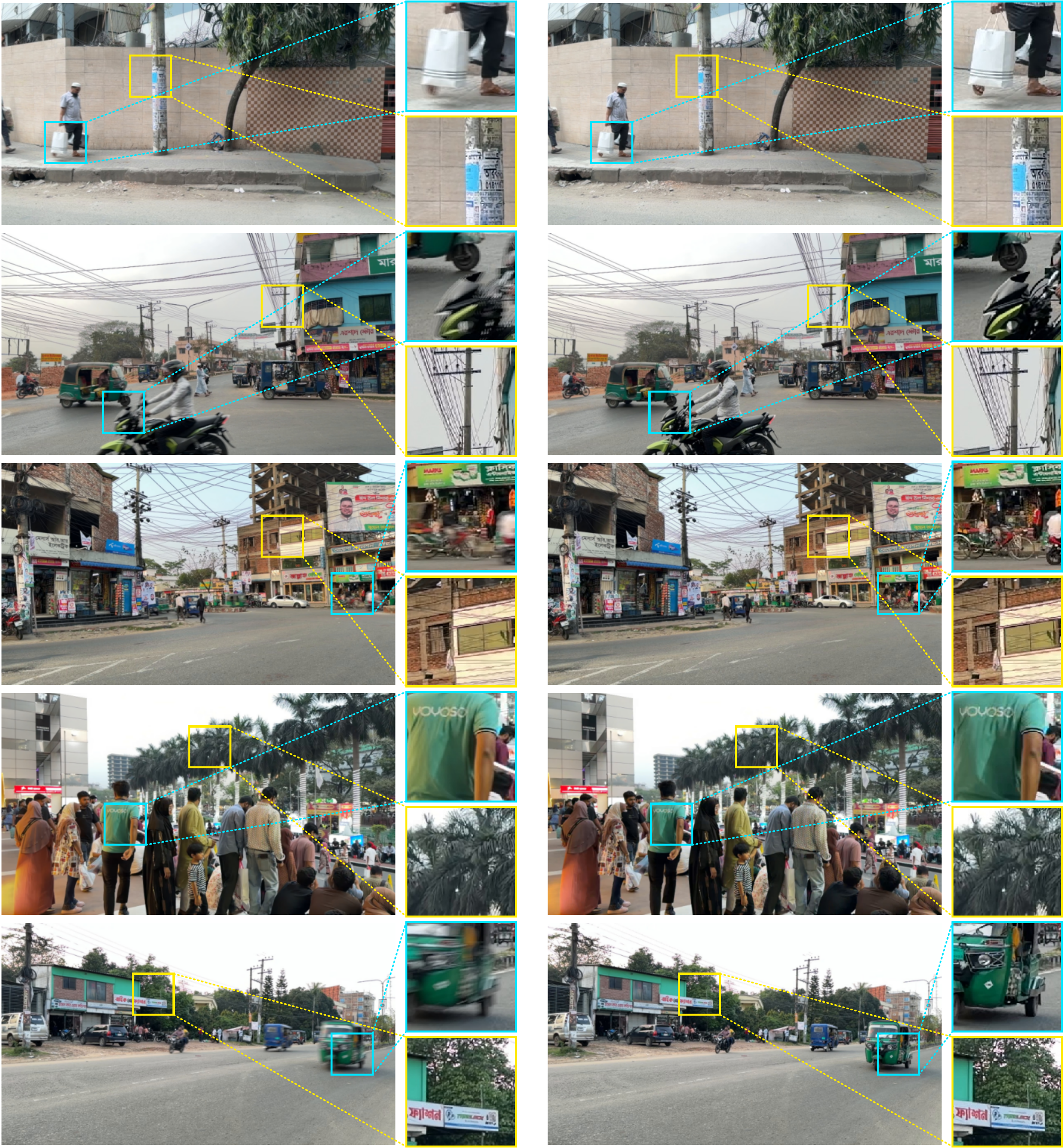}
\caption{Easy difficulty samples (PSNR $\geq$30~dB, mean optical flow 6.0~px/frame):
minimal perceptual degradation with fine details largely preserved.}
\label{fig:easy_samples}
\end{figure}

\begin{figure}[t]
\centering
\includegraphics[width=\textwidth]{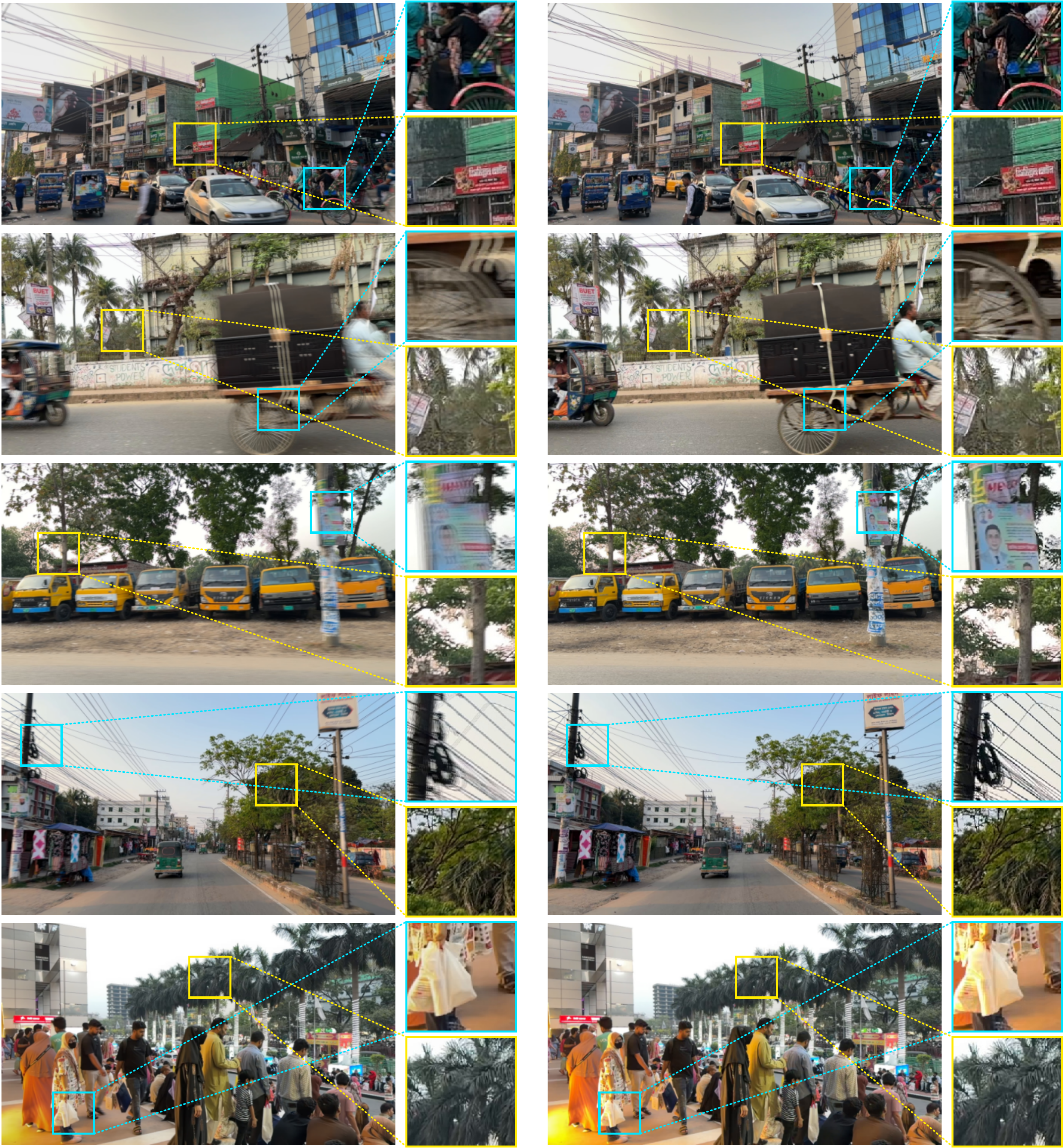}
\caption{Medium difficulty samples (PSNR 24--30~dB, mean optical flow 11.3~px/frame):
moderate blur with visible motion streaks and texture degradation.}
\label{fig:medium_samples}
\end{figure}

\begin{figure}[t]
\centering
\includegraphics[width=\textwidth]{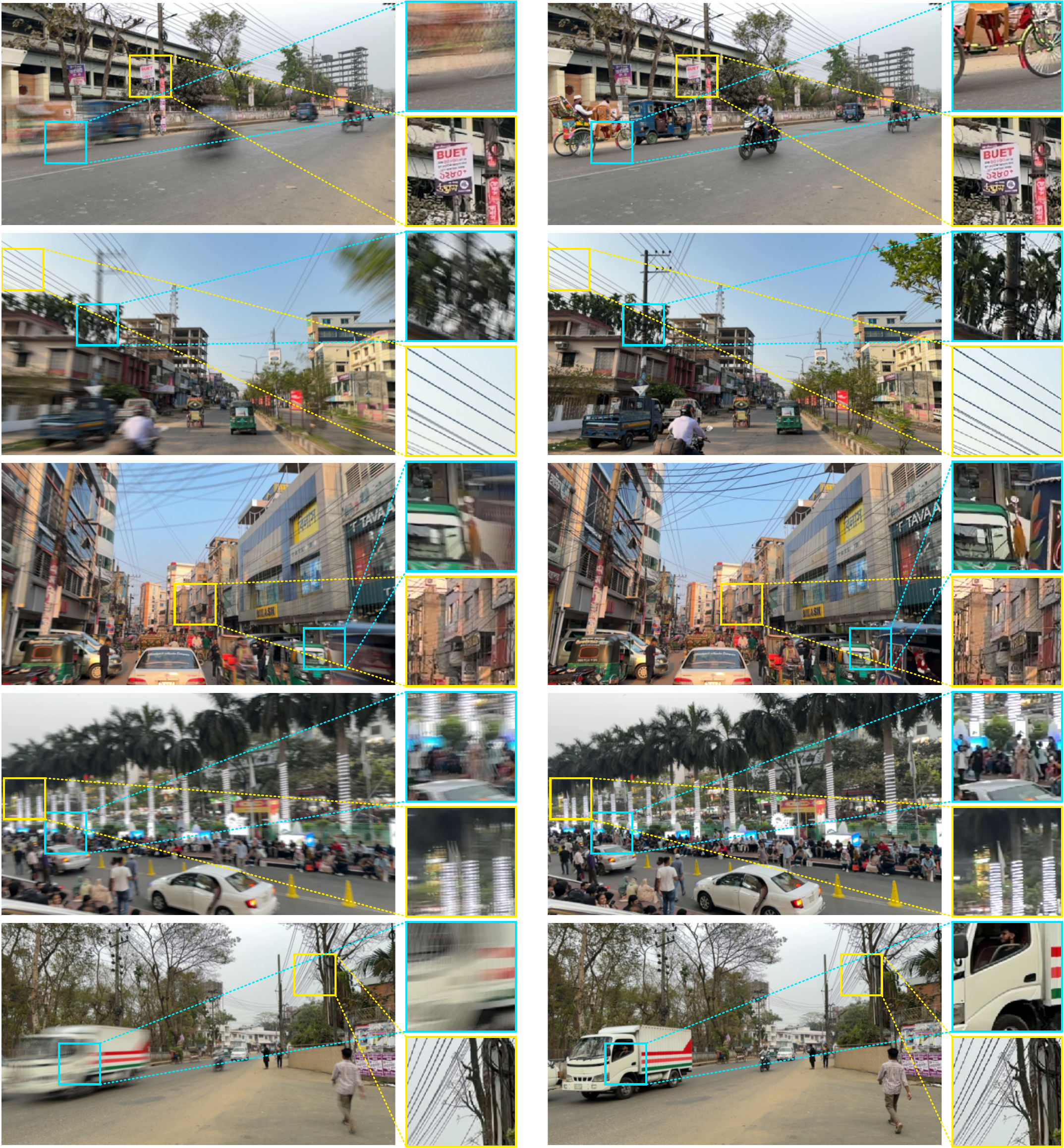}
\caption{Hard difficulty samples (PSNR $<$24~dB, mean optical flow 13.2~px/frame):
severe motion blur with substantial information loss and edge ghosting.}
\label{fig:hard_samples}
\end{figure}

Figures~\ref{fig:easy_samples}--\ref{fig:hard_samples} demonstrate content diversity
across difficulty levels, scene types, lighting conditions, and motion sources. Visual
inspection confirms difficulty stratification aligns with perceptual blur severity.

\begin{figure}[t]
\centering
\includegraphics[width=\textwidth]{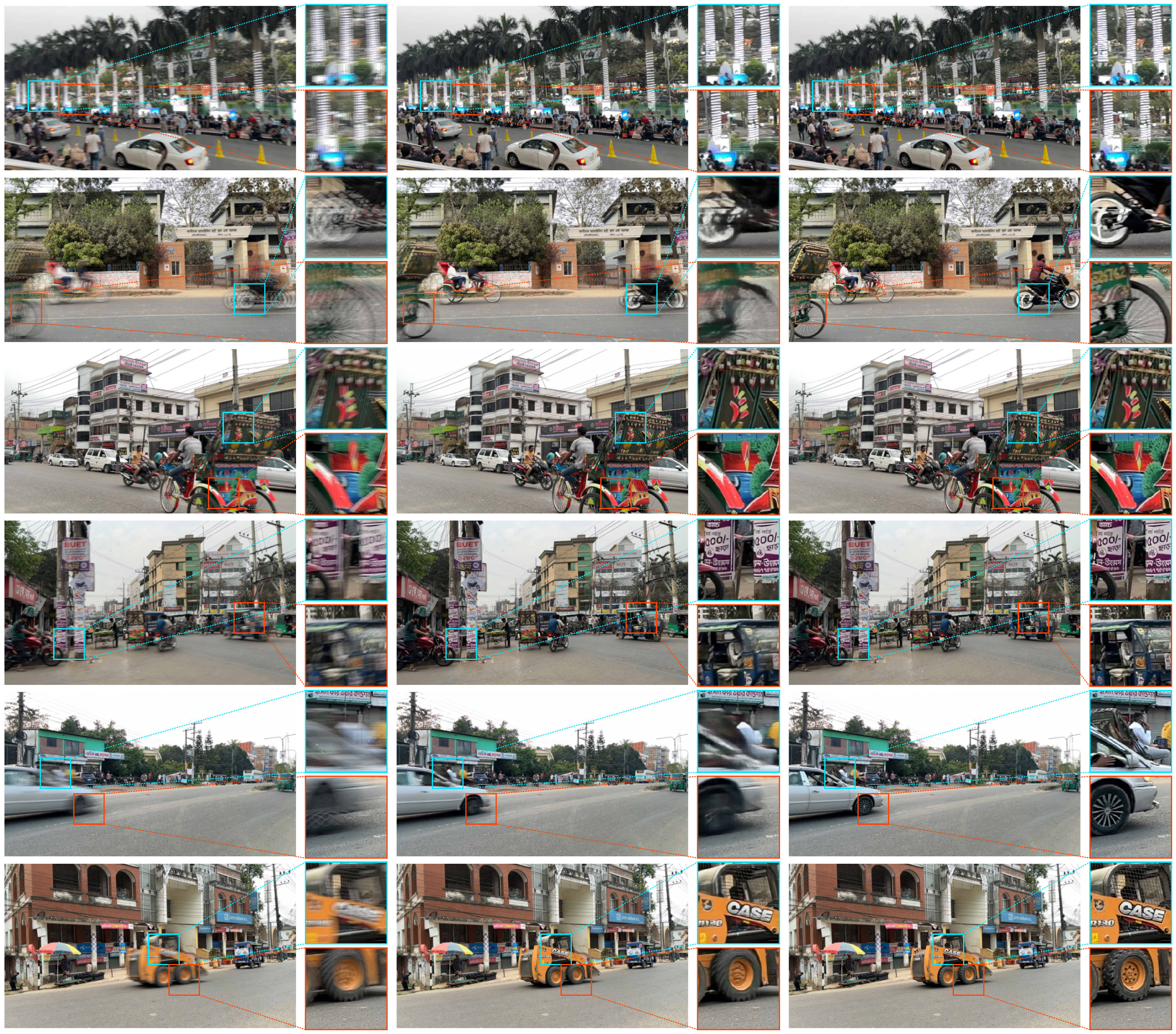}
\caption{NAFNet restoration quality across difficulty spectrum: 31.2~dB overall
(Easy 35.1~dB, Medium 31.6~dB, Hard 27.9~dB). Simple CNN architecture achieves
competitive performance.}
\label{fig:nafnet_6samples}
\end{figure}

\begin{figure}[t]
\centering
\includegraphics[width=\textwidth]{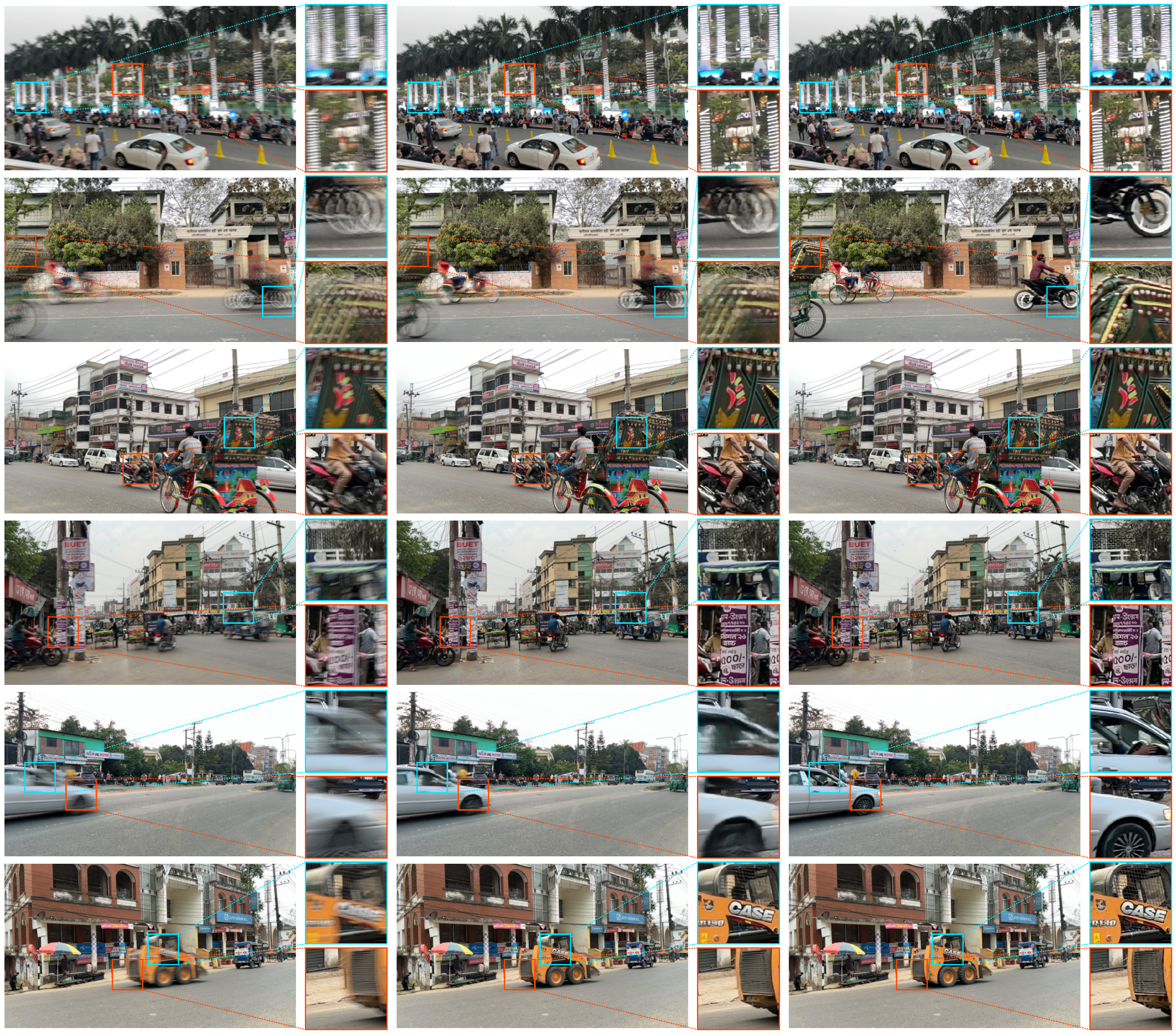}
\caption{Restormer restoration quality across difficulty spectrum: 30.9~dB overall
(Easy 34.9~dB, Medium 31.2~dB, Hard 27.5~dB). Transformer attention provides marginal
advantage.}
\label{fig:restormer_6samples}
\end{figure}

Figures~\ref{fig:nafnet_6samples}--\ref{fig:restormer_6samples} compare NAFNet
(67.9M parameters, CNN architecture) and Restormer (26.1M parameters, transformer‑based)
across six representative samples (two per difficulty tier). Both methods achieve clean
Easy restoration (PSNR $>$34~dB) but exhibit substantial Hard residual blur
(PSNR $\sim$27~dB), validating architecture‑agnostic difficulty patterns.

\begin{figure*}[t]
\centering
\includegraphics[width=\textwidth]{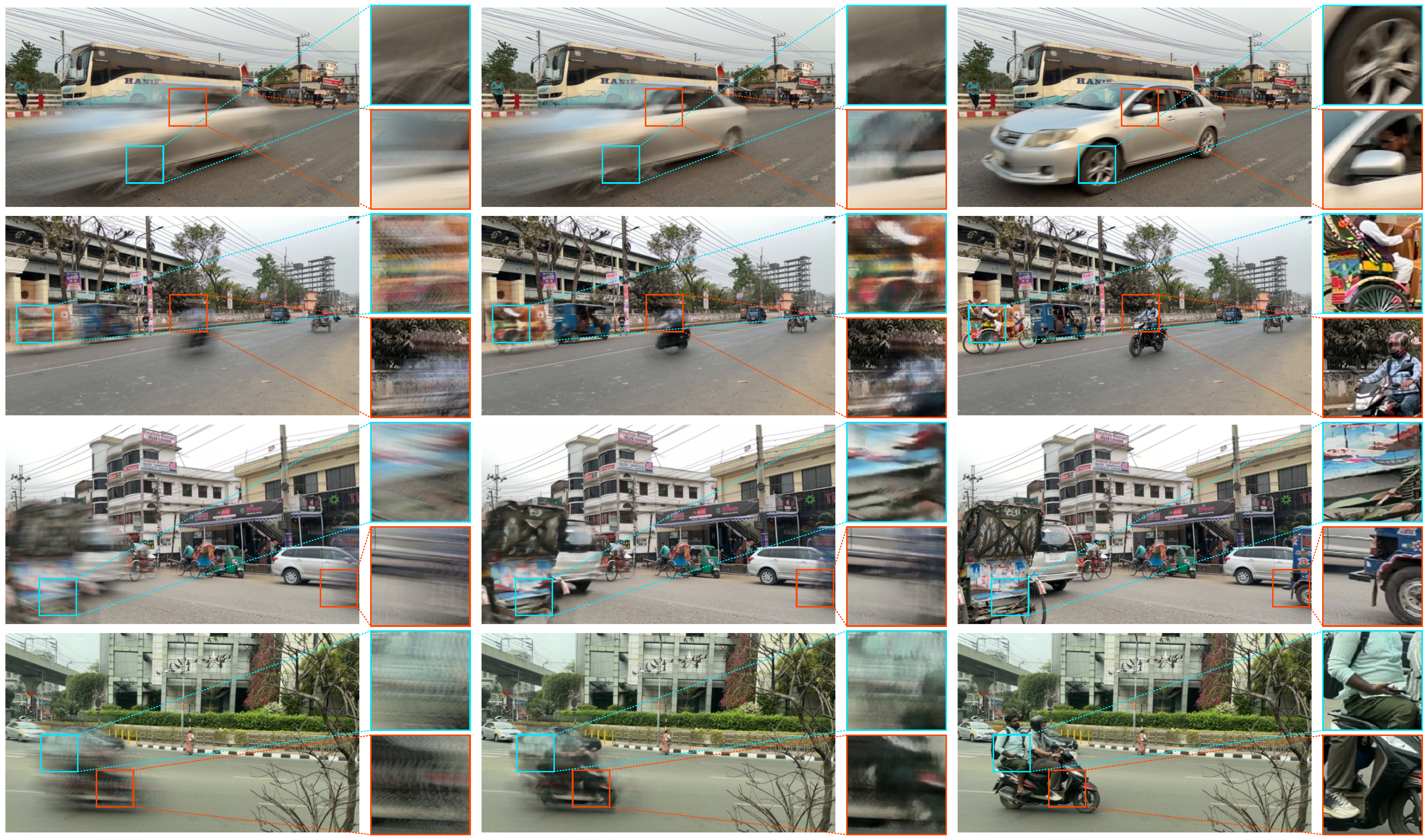}
\caption{NAFNet extreme failure cases: residual motion streaks persist despite 67.9M
parameters. Demonstrates 7.2~dB Easy‑to‑Hard performance degradation.}
\label{fig:nafnet_hard}
\vskip 10pt
\includegraphics[width=\textwidth]{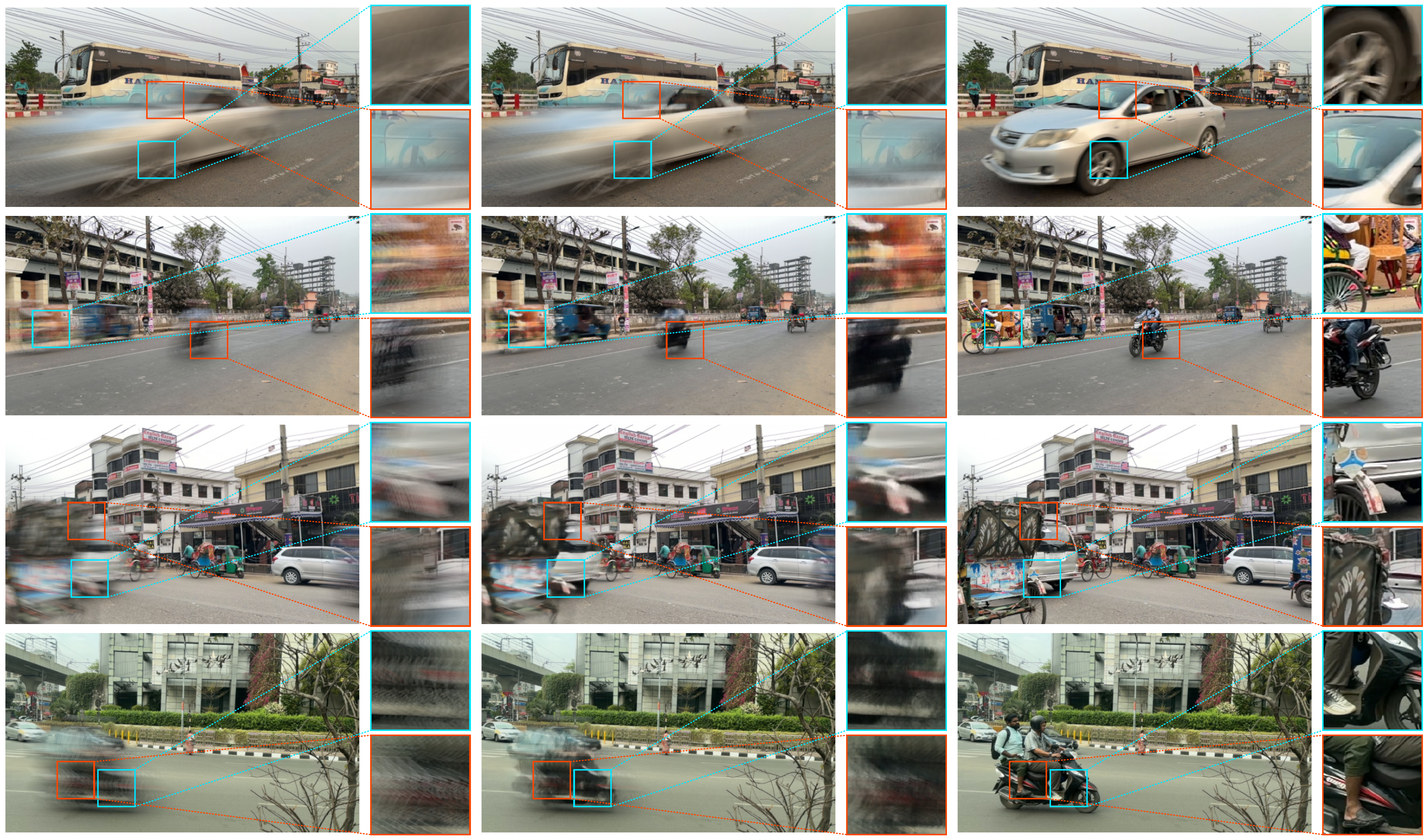}
\caption{Restormer extreme failure cases: transformer self‑attention provides minimal
advantage in severe motion scenarios. Demonstrates 7.4~dB Easy‑to‑Hard degradation.}
\label{fig:restormer_hard}
\end{figure*}

Figures~\ref{fig:nafnet_hard}--\ref{fig:restormer_hard} analyze extreme failure cases.
Both NAFNet and Restormer exhibit similar residual motion streak patterns despite
different architectural paradigms, validating that difficulty captures fundamental
physical restoration limits rather than method‑specific biases.

\section{Cross‑Dataset Validation}
\label{app:cross_validation}

\subsection{Threshold Generalization Analysis}

Applying identical PSNR thresholds (Easy $\geq$30~dB, Medium 24--30~dB, Hard $<$24~dB)
to existing benchmarks yields consistent Medium‑dominant distributions, validating that
the thresholds capture intrinsic difficulty:

\begin{itemize}[noitemsep,topsep=2pt]
\item \textbf{GoPro} test (1,111 pairs): Easy 6.6\%, Medium 55.1\%, Hard 38.3\%
\item \textbf{HIDE} test (2,025 pairs): Easy 5.1\%, Medium 39.2\%, Hard 55.8\%
\item \textbf{iPhoneBlur} (7,400 pairs): Easy 17.3\%, Medium 52.6\%, Hard 30.1\%
\end{itemize}

All datasets exhibit Medium concentration (39--55\%), confirming threshold‑based
stratification generalises across capture protocols.

\subsection{Perceptual Validation via User Study}

15 participants (computer vision researchers) rated 30 randomly sampled images
(10 per difficulty tier) on a 5‑point severity scale (1=imperceptible, 5=severe).
Results:

\begin{itemize}[noitemsep,topsep=2pt]
\item Mean ratings: Easy $2.0 \pm 0.8$, Medium $3.2 \pm 0.9$, Hard $4.1 \pm 0.7$
\item Spearman correlation (rating vs PSNR): $\rho=0.74$ ($p<0.001$)
\item ANOVA confirms significant differences between tiers ($p<0.001$)
\end{itemize}

All participants gave verbal informed consent. The study was conducted in accordance
with institutional guidelines for minimal‑risk research involving anonymised,
non‑sensitive image rating and required no formal IRB review.
These results indicate that the difficulty tiers correspond to perceptually distinct
levels of blur severity.

\section{Construction Pipeline Details}
\label{app:construction}

\subsection{Source Material and Frame Extraction}

51 source videos captured using iPhone 17 Pro at 177--240~fps (mean 208~fps),
1920$\times$1080 resolution, H.265/HEVC codec. Total approximately 741K frames.
Candidate sharp frame identification via sharpness filtering (Laplacian variance),
contrast thresholding (RMS luminance), and adaptive sampling. Final candidate pool:
10,247 high‑quality frames.

\subsection{Linearized Blur Synthesis}

Blur synthesized via $\gamma$-linearized temporal averaging (Eq.~1 in main paper).
Adaptive window selection enumerates odd sizes $W \in \{3,5,7,\ldots,21\}$, selecting
the closest to the target PSNR.

\subsection{Quality Filtering Pipeline}

Four‑stage sequential filtering removes artifacts:
\begin{enumerate}[noitemsep,topsep=2pt]
\item \textbf{Structural filter:} SSIM threshold ensures preservation (removes 823 pairs)
\item \textbf{Motion filter:} Optical flow threshold for Medium/Hard (removes 1,104 pairs)
\item \textbf{Perceptual filter:} LPIPS threshold (removes 561 pairs)
\item \textbf{Deduplication:} Perceptual hash removes near‑duplicates (removes 359 pairs)
\end{enumerate}

Final dataset: 7,400 pairs (72\% retention from 10,247 candidates).

\subsection{Train‑Test Split Protocol}

Video‑level split (75/25 ratio) prevents temporal correlation leakage. Split maintains
similar difficulty distributions: Train (Easy 17.0\%, Medium 52.8\%, Hard 30.2\%),
Test (Easy 18.3\%, Medium 51.5\%, Hard 30.2\%).

\subsection{Computational Resources}

\textbf{Synthesis:} Approximately 48 GPU‑hours on NVIDIA Tesla P100 for dataset
generation.

\textbf{Storage:} Dataset size approximately 10 GB (7,400 pairs, JPEG format).

\end{document}